# SAS+ Planning as Satisfiability

**Ruoyun Huang**                                              RUOYUN.HUANG@WUSTL.EDU
**Yixin Chen**                                                      CHEN@CSE.WUSTL.EDU
**Weixiong Zhang**                                      WEIXIONG.ZHANG@WUSTL.EDU

*Department of Computer Science and Engineering*
*Washington University in St. Louis*
*Saint Louis, Missouri, 63130, USA*

## Abstract

Planning as satisfiability is a principal approach to planning with many eminent advantages. The existing planning as satisfiability techniques usually use encodings compiled from STRIPS. We introduce a novel SAT encoding scheme (SASE) based on the SAS+ formalism. The new scheme exploits the structural information in SAS+, resulting in an encoding that is both more compact and efficient for planning. We prove the correctness of the new encoding by establishing an isomorphism between the solution plans of SASE and that of STRIPS based encodings. We further analyze the transition variables newly introduced in SASE to explain why it accommodates modern SAT solving algorithms and improves performance. We give empirical statistical results to support our analysis. We also develop a number of techniques to further reduce the encoding size of SASE, and conduct experimental studies to show the strength of each individual technique. Finally, we report extensive experimental results to demonstrate significant improvements of SASE over the state-of-the-art STRIPS based encoding schemes in terms of both time and memory efficiency.

## 1. Introduction

Planning as satisfiability (SAT) is one of the main paradigms for planning. Methods using this technique usually compile a planning problem into a sequence of SAT instances, with increasing time horizons (Kautz & Selman, 1999). Planning as satisfiability has a number of distinct characteristics that make it efficient and widely applicable. It makes use of the extensive advancement in fast SAT solvers. The SAT formulae can be extended to accommodate a variety of complex problems, such as planning with uncertainty (Castellini, Giunchiglia, & Tacchella, 2003), numerical planning (Hoffmann, Kautz, Gomes, & Selman, 2007) and temporally expressive planning (Huang, Chen, & Zhang, 2009).

A key factor for the performance of the planning as satisfiability approaches is the SAT encoding scheme, which is the way a planning problem is compiled into SAT formulae with boolean variables and clauses. As the encoding scheme has a great impact on the efficiency of SAT-based planning, developing novel and superior SAT encodings has been an active research topic. Extensive research has been done to make the SAT encoding more compact. One example of compact encoding is the lifted action representation (Kautz & Selman, 1996; Ernst, Millstein, & Weld, 1997). In this compact encoding scheme, an action is represented by a conjunction of parameters. As a result, this method mitigates the issue of blowing up encoding size. The original scheme does not guarantee the optimality on makespans. However, an improved lifted action representation that preserves optimality was proposed (Robinson, Gretton, Pham, & Sattar, 2009). A new encoding is proposed based on a relaxed parallelism semantic (Rintanen, Heljanko, & Niemelä, 2006), which also does not guarantee optimality.

All these previous enhancements are based on the conventional STRIPS formalism for planning. Recently, the SAS+ formalism (Bäckström & Nebel, 1996) attracted a lot of attention because of its rich structural information. The SAS+ formalism represents a planning problem using multi-valued state variables instead of the propositional facts in STRIPS (Bäckström & Nebel, 1996). The SAS+ formalism has been used to derive heuristics (Helmert, 2006; Helmert, Haslum, & Hoffmann, 2008), landmarks (Richter, Helmert, & Westphal, 2008), new search models (Chen, Huang, & Zhang, 2008), and strong mutual exclusion constraints (Chen, Huang, Xing, & Zhang, 2009).

In this paper, we proposed the first SAS+ based SAT encoding scheme (SASE) for classical planning. Unlike previous STRIPS based SAT encoding schemes that model only actions and facts, SASE directly models *transitions* in the SAS+ formalism. Transitions can be viewed as a high-level abstraction of actions, and there are typically significantly fewer transitions than actions in a planning task. The proposed SASE scheme describes two major classes of constraints: first the constraints between transitions and second the constraints that match actions with transitions. We theoretically and empirically studied the new SAS+ based SAT encoding and compared it against the traditional STRIPS based SAT encoding. To further improve the performance of SASE, we proposed a number of techniques to reduce encoding size by recognizing certain structures of actions and transitions.

We studied the relationship between the solution space of SASE and that of STRIPS based encoding. The results showed that the solution plans found by SATPlan06, a representative STRIPS based encoding, and by SASE are isomorphic, meaning that there is a bijective mapping between the two. Hence, we showed the equivalence between solving the STRIPS based encoding and SASE.

As an attempt to understand the performance gain of SASE, we studied how the new encoding scheme makes a SAT solving algorithm behave in a more favorable way. The study was quantified by the widely used VSIDS heuristic (Moskewicz, Madigan, Zhao, Zhang, & Malik, 2001). The transition variables that we introduced have higher frequencies in clauses, and consequently have higher VSIDS scores. The higher VSIDS scores lead to more branching on transition variables than action variables. Since the transition variables have high scores and hence stronger constraint propagation, branching more on the transition variables leads to faster SAT solving. We provided empirical evidence to support our explanation. Moreover, we introduced an indicator called the transition index, and empirically showed that there is a strong correlation between the transition index and SAT solving speedup.

Finally, we evaluated SASE on the standard benchmarks from the recent International Planning Competitions. Our results show that the new SASE encoding scheme is more efficient in both terms of time and memory usage compared to STRIPS-based encodings, and solves some large instances that the state-of-the-art STRIPS-based SAT planners fail to solve.

The paper is organized as follows. After giving some basic definitions in Section 2, we present our SAS+ based SASE encoding in Section 3 and prove its equivalence to the STRIPS based encoding in Section 4. We study why SASE works better for modern SAT solvers in Section 5. The techniques to further reduce the encoding size are presented in Section 6. We present our experimental results in Section 7. Finally, we review related works and conclude in Section 8.

## 2. Background

In this section, we first briefly introduce the STRIPS formalism and review a representative STRIPS based SAT encoding. Then, we define the SAS+ formalism, from which we develop the new SAT encoding scheme.

### 2.1 The STRIPS Formalism

The traditional STRIPS planning representation is defined over binary-valued propositional facts. A *STRIPS planning problem* is a tuple $\Psi = (\mathcal{F}, \mathcal{A}, \varphi_{\mathcal{I}}, \varphi_{\mathcal{G}})$, where:

- $\mathcal{F}$ is a set of propositional facts;

- $\mathcal{A}$ is a set of actions. Each action $a \in \mathcal{A}$ is a triple $a = (pre(a), add(a), del(a))$, where $pre(a) \subseteq \mathcal{F}$ is the set of preconditions, and $add(a) \subseteq \mathcal{F}$ and $del(a) \subseteq \mathcal{F}$ are the sets of add facts and delete facts, respectively;

- A state $\varphi \subseteq \mathcal{F}$ is a subset of facts that are assumed true. Any fact not in $\varphi$ is assumed false in this state. $\varphi_{\mathcal{I}} \subseteq \mathcal{F}$ is the initial state, and $\varphi_{\mathcal{G}} \subseteq \mathcal{F}$ is the specification of a goal state or goal states.

We define three sets of actions. We use $\text{ADD}(f)$ to denote the set of actions that have $f$ as one of their add effects, meaning $\text{ADD}(f) = \{a \mid f \in add(a)\}$. Similarly, two other action sets are $\text{DEL}(f) = \{a \mid f \in del(a)\}$ and $\text{PRE}(f) = \{a \mid f \in pre(a)\}$.

An action $a$ is applicable to a state $\varphi$ if $pre(a) \subseteq \varphi$. We use $apply(\varphi, a)$ to denote the state after applying an applicable action $a$ to $\varphi$, in which variable assignments are changed into $(\varphi \setminus del(a)) \cup add(a)$. We also write $apply(s, P)$ to denote the state after applying a set of actions $P$ in parallel, $P \subseteq \mathcal{A}$, to $s$. A set of actions $P$ is applicable to $\varphi$, when 1) each $a \in P$ is applicable to $\varphi$, and 2) there does not exist two actions $a_1, a_2 \in P$ such that $a_1$ and $a_2$ are mutually exclusive (mutex) (Blum & Furst, 1997). Two actions $a$ and $b$ are mutex at time step $t$ when one of the following three conditions holds:

- *Inconsistent effects*: $del(a) \cap add(b) \neq \emptyset$ or $del(b) \cap add(a) \neq \emptyset$.

- *Interference*: $del(a) \cap pre(b) \neq \emptyset$ or $del(b) \cap pre(a) \neq \emptyset$.

- *Competing needs*: There exist $f_1 \in pre(a)$ and $f_2 \in pre(b)$, such that $f_1$ and $f_2$ are mutex at time step $t - 1$.

Two facts $f_1$ and $f_2$ are mutex at a time step if, for all actions $a$ and $b$ such that $f_1 \in add(a)$, $f_2 \in add(b)$, $a$ and $b$ are mutex at the previous time step. We call this mutex defined on planning graphs as **P-mutex**, in order to distinguish this mutex from another notion of mutex in the next section.

**Definition 1 (Parallel solution plan).** *For a STRIPS planning problem $\Psi = (\mathcal{F}, \mathcal{A}, \varphi_{\mathcal{I}}, \varphi_{\mathcal{G}})$, a parallel solution plan is a sequence $P = \{P_1, P_2, \ldots, P_N\}$, where each $P_t \subseteq \mathcal{A}$, $t = 1, 2, \ldots, N$, is a set of actions executed at time step $t$, such that*

$$\varphi_{\mathcal{G}} \subseteq apply(\ldots apply(apply(\varphi_{\mathcal{I}}, P_1), P_2) \ldots P_N).$$

## 2.2 STRIPS Based SAT Encoding (PE)

A SAT instance is a tuple $(V, C)$, where $V$ is a set of variables and $C$ is a set of clauses. Given a SAT instance $(V, C)$, an **assignment** $\Gamma$ sets every variable $v \in V$ true or false, denoted as $\Gamma(v) = \top$ or $\Gamma(v) = \bot$. If an assignment $\Gamma$ makes every clause in $C$ to be true, then $\Gamma$ is a **solution** to $(V, C)$.

The encoding scheme by SatPlan06 (Kautz, Selman, & Hoffmann, 2006) (denoted as **PE** in the following), which is compiled from planning graphs, is a well known and extensively tested STRIPS based encoding. To facilitate the encoding, SatPlan06 introduces a dummy action $dum_f$ which has $f$ as both its precondition and add-effect. We use $\mathcal{A}^+$ to denote the set of actions when dummy actions are added, which is $A \cup \{dum_f \mid \forall f \in \mathcal{F}\}$. Unless otherwise indicated, action set $\mathrm{ADD}(f)$, $\mathrm{DEL}(f)$, and $\mathrm{PRE}(f)$ all include the corresponding dummy actions.

We denote a SatPlan06 encoding up to time step $N$ as $\mathrm{PE}(\Psi, N)$, for a given STRIPS task $\Psi = (\mathcal{F}, \mathcal{A}, \varphi_\mathcal{I}, \varphi_\mathcal{G})$. As a SAT instance, $\mathrm{PE}(\Psi, N)$ is defined by $(V, C)$, where $V = \{W_{f,t} | f \in \mathcal{F}, t \in [1, N+1]\} \cup \{W_{a,t} | a \in \mathcal{A}^+, t \in [1, N]\}$. $W_{f,t} = \top$ indicates that $f$ is true at $t$, otherwise $W_{f,t} = \bot$. The clause set $C$ includes the following types of clauses:

I. Initial state: $(\forall f, f \in \varphi_\mathcal{I})$: $W_{f,1}$;

II. Goal state: $(\forall f, f \in \varphi_\mathcal{G})$: $W_{f,N+1}$;

III. Add effect: $(\forall f \in \mathcal{F}, t \in [1, N])$: $W_{f,t+1} \rightarrow \bigvee_{\forall a, f \in add(a)} W_{a,t}$;

IV. Precondition: $(\forall a \in \mathcal{A}^+, f \in pre(a), t \in [1, N]))$: $W_{a,t} \rightarrow W_{f,t}$;

V. Mutex of actions: $(\forall a, b \in \mathcal{A}^+, t \in [1, N]$, $a$ and $b$ are mutex): $\overline{W}_{a,t} \vee \overline{W}_{b,t}$;

VI. Mutex of facts: $(\forall f, g \in \mathcal{F}, t \in [1, N+1]$, $f$ and $g$ are mutex) : $\overline{W}_{f,t} \vee \overline{W}_{g,t}$;

Clauses in class I and II enforce that the initial state is true at the first time step, and that the goal facts need to be true at the last time step, respectively. Clauses in class III specify that if a fact $f$ is true at time step $t$, then there is at least one action $a \in \mathcal{A}^+$ at time step $t-1$ that has $f$ as an add effect. Clauses of class IV specify that if an action $a$ is true at time $t$, then all its preconditions are true at time $t$. Classes V and VI specify mutex between actions and facts, respectively.

PE is one of the most typical SAT encoding schemes for STRIPS planning. It has both action variables and fact variables, and enforces the same semantics as the one defined by a planning graph. Later we will show the equivalence between our new SASE encoding and PE.

## 2.3 The SAS+ Formalism

The SAS+ formalism (Bäckström & Nebel, 1996) represents a classical planning problem by a set of multi-valued **state variables**. A planning task $\Pi$ in the SAS+ formalism is defined as a tuple $\Pi = \{\mathcal{X}, \mathcal{O}, s_\mathcal{I}, s_\mathcal{G}\}$, where

- $\mathcal{X} = \{x_1, \cdots, x_N\}$ is a set of *state variables*, each with an associated finite domain $Dom(x_i)$;

- $\mathcal{O}$ is a set of actions and each action $a \in \mathcal{O}$ is a tuple $(pre(a), eff(a))$, where $pre(a)$ and $eff(a)$ are sets of partial state variable assignments in the form of $x_i = v, v \in Dom(x_i)$;

- A state $s$ is a full assignment (a set of assignments that assigns a value to every state variable). If an assignment $(x = f)$ is in $s$, we can write $s(x) = f$. We denote $\mathcal{S}$ as the set of all states.

- $s_{\mathcal{I}} \in \mathcal{S}$ is the initial state, and $s_{\mathcal{G}}$ is a partial assignment of some state variables that define the goal. A state $s \in \mathcal{S}$ is a goal state if $s_{\mathcal{G}} \subseteq s$.

We first define what transition is. In this paper, we build constraints by recognizing that transitions are atomic elements of state transitions. Actions, cast as constraints as well in our case, act as another layer of logic flow over transitions.

**Definition 2** (**Transition**). *For a SAS+ planning task* $\Pi = \{\mathcal{X}, \mathcal{O}, s_{\mathcal{I}}, s_{\mathcal{G}}\}$, *given a state variable* $x \in \mathcal{X}$, *a transition is a re-assignment of* $x$ *from value* $f$ *to* $g$, $f, g \in Dom(x)$, *written as* $\delta^x_{f \to g}$, *or from an unknown value to* $g$, *written as* $\delta^x_{* \to g}$. *We may also simplify the notation of* $\delta^x_{f \to g}$ *as* $\delta_{f \to g}$ *or* $\delta$, *when there is no confusion.*

Transitions in a SAS+ planning task can be classified into three categories.

- Transitions of the form $\delta^x_{f \to g}$ are called **regular**. A regular transition $\delta^x_{f \to g}$ is applicable to a state $s$, iff $s(x) = f$. Let $s' = apply(s, \delta^x_{f \to g})$ be the state after applying transition $\delta$ to state $s$, we have $s'(x) = g$.

- Transitions of the form $\delta^x_{f \to f}$ are called **prevailing**. A prevailing transition $\delta^x_{f \to f}$ is applicable to a state $s$ iff $s(x) = f$, and $apply(s, \delta^x_{f \to f}) = s$.

- Transitions of the form $\delta^x_{* \to g}$ are called **mechanical**. A mechanical transition $\delta^x_{* \to g}$ can be applied to an arbitrary state $s$, and the result of $apply(s, \delta^x_{* \to g})$ is a state $s'$ with $s'(x) = g$.

A transition is applicable at a state $s$ only in the above three cases. For each action $a$, we denote its **transition set** as $Trans(a)$, which includes: all regular transitions $\delta^x_{f \to g}$ such that $(x = f) \in pre(a)$ and $(x = g) \in \textit{eff}(a)$, all prevailing transitions $\delta^x_{f \to f}$ such that $(x = f) \in pre(a)$, and all mechanical transitions $\delta^x_{* \to g}$ such that $(x = g) \in \textit{eff}(a)$. Given a transition $\delta$, we use $A(\delta)$ to denote the set of actions $a$ such that $\delta \in Trans(a)$. We call $A(\delta)$ the **supporting action set** of $\delta$.

For a state variable $x$, we introduce $\mathcal{T}(x) = \{\delta^x_{f \to g}\} \cup \{\delta^x_{f \to f}\} \cup \{\delta^x_{* \to g}\}$, for all $f, g \in Dom(x)$, which is the set of all transitions that affect $x$. We also define $\mathcal{T}$ as the union of $\mathcal{T}(x)$, $\forall x \in \mathcal{X}$. $\mathcal{T}$ is the set of all transitions. We also use $R(x) = \{\delta^x_{f \to f} \mid \forall f, f \in Dom(x)\}$ to denote the set of all prevailing transitions related to $x$, and $R$ the union of $R(x)$ for all $x \in \mathcal{X}$.

**Definition 3** (**Transition Mutex**). *For a SAS+ planning task, two different transitions* $\delta_1$ *and* $\delta_2$ *are mutually exclusive iff there exists a state variable* $x \in \mathcal{X}$ *such that* $\delta_1, \delta_2 \in \mathcal{T}(x)$, *and one of the following holds:*

1. *Neither* $\delta_1$ *nor* $\delta_2$ *is a mechanical transition.*

2. *At least one of* $\delta_1$ *and* $\delta_2$ *is a mechanical transition, with* $\delta_1$ *and* $\delta_2$ *transit to different values.*

A set of transitions $T$ is applicable to a state $s$ when 1) every transition $\delta \in T$ is applicable to $s$, and 2) there do not exist two transitions $\delta_1, \delta_2 \in T$ such that $\delta_1$ and $\delta_2$ are mutually exclusive. When $T$ is applicable to $s$, we write $apply(s, T)$ to denote the state after applying all transitions in $T$ to $s$ in an arbitrary order.

**Definition 4** **(Transition Plan)**. *A transition plan is a sequence of $\{T_1, T_2, \ldots, T_N\}$, where each $T_t$, $t \in [1, N]$, is a set of transitions executed at time step $t$, such that*

$$s_\mathcal{G} \subseteq apply(\ldots apply(apply(s_\mathcal{I}, T_1), T_2) \ldots T_N).$$

In a SAS+ planning task, for a given state $s$ and an action $a$, when all variable assignments in $pre(a)$ match the assignments in $s$, $a$ is *applicable* in state $s$. We use $apply(s, a)$ to denote the state after applying $a$ to $s$, in which variable assignments are changed according to *eff*$(a)$.

**Definition 5** **(S-Mutex)**. *For a SAS+ planning task $\Pi = \{\mathcal{X}, \mathcal{O}, s_\mathcal{I}, s_\mathcal{G}\}$, two actions $a_1, a_2 \in \mathcal{O}$ are S-mutex iff either of the following holds:*

1. *There exists a transition $\delta$, such that $\delta$ is not prevailing ($\delta \notin R$), and $\delta \in Trans(a_1)$ and $\delta \in Trans(a_2)$. Actions $a_1$ and $a_2$ in this case deletes each other's precondition.*

2. *There exist two transitions $\delta$ and $\delta'$ such that they are mutually exclusive to each other and $\delta \in Trans(a_1)$ and $\delta' \in Trans(a_2)$.*

We named this mutex in SAS+ planning S-mutex to distinguish it from the P-mutex defined in STRIPS planning. We will show in Section 4 that these two types of mutual exclusions are equivalent. Therefore, in this paper we in general use the single term *mutual exclusion (mutex)* for both, unless otherwise indicated.

For a SAS+ planning task, we write $apply(s, P)$ to denote the state after applying a set of actions $P$, $P \subseteq \mathcal{O}$, to $s$. A set of actions $P$ is applicable to $s$ when 1) each $a \in P$ is applicable to $s$, and 2) there are no two actions $a_1, a_2 \in P$ such that $a_1$ and $a_2$ are S-mutex.

**Definition 6** **(Action Plan)**. *For a SAS+ task, an action plan is a sequence $P = \{P_1, \ldots, P_N\}$, where each $P_t$, $t \in [1, N]$, is a set of actions executed at time step $t$ such that*

$$s_\mathcal{G} \subseteq apply(\ldots apply(apply(s_\mathcal{I}, P_1), P_2) \ldots P_N).$$

The definition of an action plan for SAS+ planning is essentially the same as that for STRIPS planning (Definition 1). The relation between transition plan and action plan is the key to the new encoding scheme introduced in this paper. There always exists a unique transition plan for a valid action plan. In contrast, given a transition plan, there may not be a corresponding action plan; or there could be multiple corresponding action plans.

**Definition 7** **(Step Optimal Plan)**. *For a SAS+ planning task, a step optimal plan is an action plan $P = \{P_1, \ldots, P_N\}$ with the minimum $N$.*

It is worth noting that there are a few different optimization metrics in classical planning research, including step optimality (Definition 7), the number of actions and the total action cost. The criteria used in recent IPC competitions (The 6th Int'l Planning Competition, 2008; The 7th Int'l Planning Competition, 2011) is the total action cost. While step optimality is a widely used criterion, action cost is a more realistic criterion in many domains such as those involving numerical resources. Nevertheless, action cost assumes plans to be sequential. In other words, it does not consider concurrency between actions, which is a limitation in many cases.

The step optimality was introduced in GraphPlan (Blum & Furst, 1997), and became more popular when planning graph analysis was used in several planning systems (Kautz & Selman, 1999; Hoffmann & Nebel, 2001; Do & Kambhampati, 2000). Step optimality takes the concurrency between actions into consideration, although it assumes a same unit duration for all actions. The planning methods for step optimality, in particular SAT-based planners, can potentially be made useful for other optimization metrics, which is a topic of our future work.

## 3. SAS+ Based SAT Encoding (SASE)

We now introduce our new encoding for SAS+ planning tasks, denoted as SASE. We use the same search framework as SatPlan: start with a small number of time steps $N$ and increase $N$ by one after each step until a satisfiable solution is found. For a given $N$, we encode a planning task as a SAT instance which can be solved by a SAT solver. A SASE instance includes two types of binary variables:

1. Transition variables: $U_{\delta,t}$, $\forall \delta \in \mathcal{T}$ and $t \in [1, N]$, which may also be written as $U_{x,f,g,t}$ when $\delta$ is explicitly $\delta^x_{f \to g}$;

2. Action variables: $U_{a,t}$, $\forall a \in \mathcal{O}$ and $t \in [1, N]$.

As to constraints, SASE has eight classes of clauses for a SAS+ planning task. In the following, we define each class for every time step $t \in [1, N]$ unless otherwise indicated.

A. Initial state: $\forall x, s_{\mathcal{I}}(x) = f, \bigvee_{\forall \delta_{f \to g} \in \mathcal{T}(x)} U_{x,f,g,1}$;

B. Goal: $\forall x, s_{\mathcal{G}}(x) = g, \bigvee_{\forall \delta_{f \to g} \in \mathcal{T}(x)} U_{x,f,g,N}$;

C. Progression: $\forall \delta^x_{h \to f} \in \mathcal{T}$ and $t \in [1, N-1], U_{x,h,f,t} \to \bigvee_{\forall \delta^x_{f \to g} \in \mathcal{T}(x)} U_{x,f,g,t+1}$;

D. Regression: $\forall \delta^x_{f \to g} \in \mathcal{T}$ and $t \in [2, N], U_{x,f,g,t} \to \bigvee_{\forall \delta^x_{f' \to f} \in \mathcal{T}(x)} U_{x,f',f,t-1}$;

E. Transition mutex: $\forall \delta_1 \forall \delta_2$ such that $\delta_1$ and $\delta_2$ are transition mutex, $\overline{U}_{\delta_1,t} \vee \overline{U}_{\delta_2,t}$;

F. Composition of actions: $\forall a \in \mathcal{O}, U_{a,t} \to \bigwedge_{\forall \delta \in Trans(a)} U_{\delta,t}$;

G. Action existence: $\forall \delta \in \mathcal{T} \setminus R, U_{\delta,t} \to \bigvee_{\forall a, \delta \in Trans(a)} U_{a,t}$;

H. Action mutex: $\forall a_1 \forall a_2$ such that $\exists \delta, \delta \in T(a_1) \cap T(a_2)$ and $\delta \notin R, \overline{U}_{a_1,t} \vee \overline{U}_{a_2,t}$;

Clauses in classes C and D specify and restrict how transitions change over time steps. Clauses in class E enforce that at most one related transition can be true for each state variable at each time step. Clauses in classes F and G together encode how actions are composed to match the transitions. Clauses in class H enforce the mutual exclusions between actions.

Note that there are essential differences between transition variables in SASE and fact variables in PE. In terms of semantics, a transition variable at time step $n$ in SASE is equivalent to the conjunction of two fact variables in PE, at time step $n$ and $n + 1$, respectively. Nevertheless, fact variables are not able to enforce a transition plan as transition variables do. This is because transition variables not only imply the values of multi-valued variables, but also enforce how these values propagate over time steps.

In addition, transition variables are different from action variables regarding their roles in SAT solving. This is because in SASE, action variables only exist in the constraints for transition-action matching, but not in the constraints between time steps. Transition variables exist in both. Thus transition variables appear more frequently in a SAT instance. The inclusion of those high-frequency variables can help the SAT solvers through the VSIDS rule for variable branching. We shall discuss this issue further and provide an empirical study, in Section 5.

We now show how SASE works using an example. Consider a planning task with two multi-valued variables $x$ and $y$, where $Dom(x) = \{f, g, h\}$ and $Dom(y) = \{d, e\}$. There are three actions $a1 = \{\delta^x_{f \to g}, \delta^y_{d \to e}\}$, $a2 = \{\delta^x_{f \to g}, \delta^y_{e \to d}\}$ and $a3 = \{\delta^x_{g \to h}, \delta^y_{e \to d}\}$. The initial state is $\{x = f, y = d\}$ and the goal state is $\{x = h, y = d\}$. One solution to this instance is a plan of two actions: $a1$ at time step 1 and then $a3$ at time step 2.

In the following we list the constraints between transitions and actions, namely those specified in classes F and G. The clauses in other classes are self-explanatory. In particular, here we only list the variables and clauses for time step 1, because these constraints all repeat for time step 2. The transition variables at time step 1 are { $U_{x,f,g,1}, U_{x,f,f,1}, U_{x,g,h,1}, U_{x,g,g,1}, U_{x,h,h,1}, U_{y,d,d,1}, U_{y,e,e,1}$, $U_{y,e,d,1}, U_{y,d,e,1}$ }, and they repeat for time step 2. The action variables at time step 1 are { $U_{a1,1}$, $U_{a2,1}, U_{a3,1}$}, and they repeat for time step 2.

The clauses in class F are: $\overline{U}_{a1,1} \vee U_{x,f,g,1}$, $\overline{U}_{a1,1} \vee U_{x,d,e,1}$, $\overline{U}_{a2,1} \vee U_{x,f,g,1}$, $\overline{U}_{a2,1} \vee U_{y,e,d,1}$, $\overline{U}_{a3,1} \vee U_{x,g,h,1}$ and $\overline{U}_{a3,1} \vee U_{x,e,d,1}$. The clauses in class G are $\overline{U}_{x,f,g,1} \vee U_{a1,1}$, $\overline{U}_{a2,1}$, $\overline{U}_{x,g,h,1} \vee U_{a3,1}$, $\overline{U}_{y,d,e,1} \vee U_{a1,1}$, and $\overline{U}_{y,e,d,1} \vee U_{a2,1} \vee U_{a3,1}$.

The solution, in terms of actions, has action variables $U_{a1,1}$ and $U_{a3,2}$ to be $\top$, and all other action variables $\bot$. In addition, the corresponding transition plan has the following transition variables $\top$: $\{U_{x,f,g,1}, U_{x,g,h,2}, U_{y,d,e,1}, U_{y,e,d,2}\}$, while all other transition variables are false.

As mentioned above, although there are often multiple transition plans, a transition plan may not correspond to a valid action plan. In this particular example, there are several different transition plans that satisfy the initial and the goal states, but some of them do not have a corresponding action plan. For example, suppose transition variables $\{U_{x,f,g,1}, U_{x,g,h,2}, U_{y,d,d,1}, U_{y,d,d,2}\}$ to be true. This qualifies as a transition plan, because the goals are achieved. This transition plan however does not lead to a valid action plan.

## 4. Correctness of SAS+ Based Encoding

It is important to prove the correctness of the proposed encoding. We achieve this by proving that SASE for SAS+ planning has the same solution space as that of PE used in STRIPS planning. More specifically, we show that, for a given planning task and a given time step $N$, the SAT instance from SASE is satisfiable if and only if the SAT instance from PE is satisfiable. Here, we assume the correctness of the PE encoding, in SatPlan06 (Kautz et al., 2006) for STRIPS planning.

PE$(\Psi, N)$ denotes the PE formula that corresponds to the N-step planning problem. SASE$(\Pi, N)$ gives the formula in the case of SASE encoding of the equivalent SAS+ problem.

### 4.1 Solution Structure of STRIPS Based Encoding

In this section, we study a few properties of the solutions in a STRIPS based encoding. These properties provide some key insights for establishing the relationship between PE and SASE encodings.

**Lemma 1** *Given a STRIPS task* $\Psi = (\mathcal{F}, \mathcal{A}, \varphi_{\mathcal{I}}, \varphi_{\mathcal{G}})$, *a time step* $N$, *and its* PE *SAT instance* $\mathrm{PE}(\Psi, N) = (V, C)$, *suppose there is a satisfiable solution denoted as* $\Gamma$, *a fact* $f \in \mathcal{F}$, *and* $t \in [1, N]$ *such that: 1)* $\Gamma(W_{dum_f, t}) = \perp$, *2)* $\Gamma(W_{f,t}) = \top$, *and 3)* $\forall a \in \mathrm{DEL}(f)$, $\Gamma(W_{a,t}) = \perp$, *then we can construct an alternative solution* $\Gamma'$ *to* $\mathrm{PE}(\Psi, N)$ *as follows:*

$$\Gamma'(v) = \begin{cases} \top, \ v = W_{dum_f, t}, v \in V \\ \Gamma(v), \ v \neq W_{dum_f, t}, v \in V \end{cases} \tag{1}$$

**Proof** This can be proved by showing that $\Gamma'$ satisfies every individual clause in $C$. See Appendix A for more details. □

**Lemma 2** *Given a STRIPS task* $\Psi = (\mathcal{F}, \mathcal{A}, \varphi_{\mathcal{I}}, \varphi_{\mathcal{G}})$, *a time step* $N$, *and its* PE *SAT instance* $\mathrm{PE}(\Psi, N) = (V, C)$, *suppose there is a satisfiable solution denoted as* $\Gamma$, *a fact* $f \in \mathcal{F}$, *and* $t \in [1, N]$ *such that: 1)* $\Gamma(W_{f,t}) = \perp$, *2) there exists an action* $a \in \mathrm{ADD}(f)$ *such that* $\Gamma(W_{a,t-1}) = \top$, *then we can construct an alternative solution* $\Gamma'$ *to* $\mathrm{PE}(\Psi, N)$ *as follows:*

$$\Gamma'(v) = \begin{cases} \top, \ v = W_{f,t}, v \in V \\ \Gamma(v), \ v \neq W_{f,t}, v \in V \end{cases} \tag{2}$$

**Proof** We can prove this by showing that $\Gamma'$ makes every individual clause (three types) in $C$ to be true. See Appendix A for more details. □

Lemmas 1 and 2 show that under certain conditions, some dummy action variables and fact variables in PE are *free variables*. They can be set to be either true or false while the SAT instance remains satisfied. Although we can manipulate these free variables to construct an alternative solution $\Gamma'$ from a given solution $\Gamma$, both $\Gamma$ and $\Gamma'$ refer to the same STRIPS plan, because there is no change to any action variable. This leads to an important insight concerning the solutions of PE: a solution plan to a STRIPS planning problem $\Psi$ may correspond to multiple solutions to $\mathrm{PE}(\Psi, N)$.

**Proposition 1** *Given a STRIPS task* $\Psi = (\mathcal{F}, \mathcal{A}, \varphi_{\mathcal{I}}, \varphi_{\mathcal{G}})$, *a time step* $N$, *and its* PE *SAT instance* $\mathrm{PE}(\Psi, N) = (V, C)$, *those clauses that define competing needs mutex and fact mutex can be inferred from other clauses in* $\mathrm{PE}(\Psi, N)$.

These mutexes are implied by the PE formula, thus by the completeness of resolution, Proposition 1 is true. Proposition 1 implies that when encoding a STRIPS task, it is not necessary to encode fact mutex and competing needs action mutex, as they are implied by other clauses. Therefore, while considering the completeness and correctness of PE, we can ignore these redundant clauses. Analysis with a similar conclusion can be found in the literature (Sideris & Dimopoulos, 2010), although different approaches are used.

## 4.2 Equivalence of STRIPS and SAS+ Based Encodings

A classical planning problem can be represented by both STRIPS and SAS+ formalisms that give rise to the same set of solutions. Given a STRIPS task $\Psi = (\mathcal{F}, \mathcal{A}, \varphi_{\mathcal{I}}, \varphi_{\mathcal{G}})$ and its equivalent SAS+ planning task $\Pi = (\mathcal{X}, \mathcal{O}, s_{\mathcal{I}}, s_{\mathcal{G}})$, the following isomorphisms (bijective mappings) exist:

- $\phi_f : \mathcal{F} \to \prod_{\mathcal{X}} Dom(X)$ (a binary STRIPS fact corresponds to an variable assignment in SAS+);

- $\phi_a : \mathcal{A} \to \mathcal{O}$ (a STRIPS action corresponds to a SAS+ action);

- $\phi_i : \varphi_{\mathcal{I}} \to s_{\mathcal{I}}$ (can be derived from $\phi_f$);

- $\phi_g : \varphi_{\mathcal{G}} \to s_{\mathcal{G}}$ (can be derived from $\phi_f$).

Furthermore, since both formalisms represent the same planning task, these mappings preserve the relations between actions and facts. For example, if $f \in pre(a)$ where $f \in \mathcal{F}$ and $a \in \mathcal{A}$ in a STRIPS formalism, we have $\phi_f(f) \in pre(\phi_a(a))$ in the SAS+ formalism.

First, we show that the parallelism semantics enforced by S-mutex in SAS+ is equivalent to that of P-mutex in STRIPS.

**Lemma 3** *Given a SAS+ planning task $\Pi = (\mathcal{X}, \mathcal{O}, s_{\mathcal{I}}, s_{\mathcal{G}})$ and its equivalent STRIPS task $\Psi = (\mathcal{F}, \mathcal{A}, \varphi_{\mathcal{I}}, \varphi_{\mathcal{G}})$, suppose we have actions $a, b \in \mathcal{O}$, and their equivalent actions $a', b' \in \mathcal{A}$ (i.e. $a = \phi_a(a')$ and $b = \phi_a(b')$), $a$ and $b$ are S-mutex iff $a'$ and $b'$ are P-mutex.*

**Proof**   We can prove this by showing in both directions that given two actions with one type of mutex, there is also mutux of the other type. Details are in Appendix A.   □

Lemma 3 connects P-mutex and S-mutex. Based on that we can construct the relations between the encodings, which are used in the proofs of both Theorems 1 and 2, respectively.

**Theorem 1** *Given a STRIPS task $\Psi$ and a SAS+ task $\Pi$ that are equivalent, for a time step bound $N$, if $\mathrm{PE}(\Psi, N)$ is satisfiable, $\mathrm{SASE}(\Pi, N)$ is also satisfiable.*

**Proof**   We prove this theorem by construction. Suppose we know the solution to PE, we prove that we can construct a solution to SASE accordingly. Details are in Appendix A.   □

**Theorem 2** *Given a STRIPS task $\Psi$ and a SAS+ task $\Pi$ that are equivalent, for a time step bound $N$, if $\mathrm{SASE}(\Pi, N)$ is satisfiable, $\mathrm{PE}(\Psi, N)$ is also satisfiable.*

**Proof**   This can be proved by using the same technique used for Theorem 1. See Appendix A for more details.   □

From Theorems 1 and 2, we reach the following conclusion.

**Theorem 3** *A classical planning problem is solvable by the $\mathrm{PE}$ encoding if and only if it is solvable by the $\mathrm{SASE}$ encoding. Further, for solvable problems, the solution plans found by the two encodings have the same, optimal makespan.*

Theorem 3 reveals that a planning solution can be found by SASE if and only if it can be found by PE. In terms of SAT solutions, the proofs show that there is an epimorphism (a surjective mapping) between the solutions to PE and the solutions to SASE. That is, multiple SAT solutions in PE map to one SAT solution in SASE and every SAT solution in SASE is mapped from at least one SAT solution in PE. This is due to the existence of *free variables* in the PE encoding. One solution in SASE corresponds to a group of solutions in PE with the same assignments to real action variables but different assignments to the free variables.

## 5. SAT Solving Efficiency on Different Encodings

In Section 4, we showed that PE and SASE are semantically equivalent in that they have the same solution space. In this section, we study what makes them different regarding SAT solving efficiency. In particular, we want to understand how PE and SASE make a SAT solver behave differently on the same planning task. Modern SAT solvers, which nowadays all employ many sophisticated techniques, are too complicated to be characterized by simple models. In general, it is difficult to accurately estimate the time that a SAT solver needs to solve a SAT instance. In this section, we provide an explanation of why the SAT encodings from SASE are more efficient for SAT solvers to solve than the SAT encodings from PE, and provide empirical evidence to support this explanation.

In this section, we first discuss SASE's problem structure in Section 5.1 and the reason that the widely used SAT solving heuristic VSIDS (Moskewicz et al., 2001) works better on SASE encodings. The idea of VSIDS is to select those variables that appear frequently in the original and learnt clauses, since they lead to stronger constraint propagation and space pruning. If we order all the variables by their VSIDS scores, a large population of the top-ranked transition variables introduced in SASE have higher VSIDS scores than the top-ranked action variables. As a result, those top-ranked transition variables are selected more often and provide stronger constraint propagation, speeding up SAT solving.

We study the significance of transition variables, and explain why they make search more efficient. In Section 5.2, we present a comparison on transition variables versus action variables. In Section 5.3, we show how often transition variables are chosen as decision variables, which is a direct evidence of transition variables' significance.

Finally, in Section 5.4 we empirically define a significance index of transition variables. This index measures the significance of transition variables within the context of the VSDIS heuristic, and correlates with the speedup in SAT solving. All analysis in this section uses SatPlan06 as the baseline.

### 5.1 VSIDS Heuristic in SAT Solving

The SAT solvers we use are based on the Conflict Driven Clause Learning framework. A *decision variable* refers to the one selected as the next variable for branching. Once a decision variable is chosen, more variables could be fixed by unit propagation. The ordering of decision variables significantly affects the problem solving efficiency. Most existing complete SAT algorithms use variants of the VSIDS heuristic (Moskewicz et al., 2001) as their variable ordering strategy.

The VSIDS heuristic essentially evaluates a variable by using the Exponential Moving Average (EMA) of the number of times (**frequency**) it appears in all the clauses. This frequency value keeps changing because of the learnt clauses. Therefore, VSIDS uses a smoothing scheme which periodically scales down the scores of all variables by a constant in order to reflect the importance of recent changes in frequencies. The variable that occurs most frequently usually has a higher value, thus also a higher chance of being chosen as a decision variable. A random decision is made if there is a tie. Thus, variables associated with more recent conflict clauses have higher priorities. We first consider the frequency in the original clauses only. Then we investigate further by taking the periodic update into consideration.

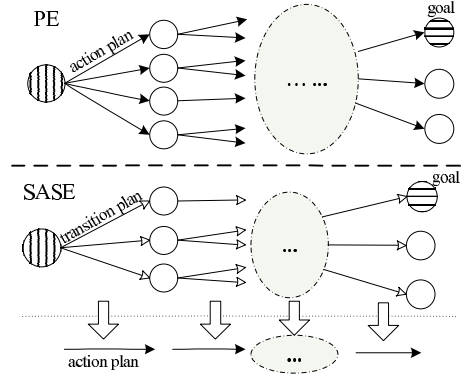

Figure 1: Illustration of how the search spaces of two encoding schemes differ from each other.

Given the fact that the frequency is used as the main measurement, VSIDS will be most effective when the difference between variables' frequencies are large and there are some variables with high frequencies. If all variables have the same frequency, then picking decision variables will be purely random. Further, variables with high frequencies are desirable since they lead to stronger constraint propagation.

The major difference between the SAT instances in PE and SASE is that in the latter encoding, actions are not responsible for the constraint propagation across time steps. Figure 1 illustrates their difference. In SASE, the SAT instance can be conceptually reduced to the following search problem of two hierarchies.

- At the top level, we search for a transition plan as defined in Definition 4. This amounts to finding a set of transitions for each time step $t$ (corresponding to all $\delta$ such that $U_{\delta,t}$ is set to $\top$), so that they satisfy the clauses in classes A-E of the SASE encoding.

- At the lower level, we try to find an action plan that satisfies the transition plan. In other words, for a given transition plan that satisfies clauses in classes A-E, we try to find an action plan satisfying clauses in classes F-H.

## 5.2 Transition Variables versus Action Variables

Let us first formulate how the frequency of a variable is measured. Given a SAT instance $(V, C)$, for each variable $v \in V$, we define a function $h(v)$ to indicate the frequency of $v$. That is, $h(v)$ is the number of clauses that $v$ appears in. We sort all variables in $V$ by their $h$ values in a descending order, and study the variables above a certain $h$ value.

**Definition 8** **(High $h$ Value Variable Set)**. *Given a SAT instance $(V, C)$, for a $h' \in \{h(v) \mid v \in V\}$, we denote $V(h')$ as the set of all variables $v \in V$ such that $h(v) \geq h'$.*

We further define *percentile* and *top variable set* to quantify our analysis. Instead of a specific $h()$ value, we use a percentage of all $h$ values to make the analysis comparable across instances.

**Definition 9  (Percentile).** *Given a SAT instance $(V, C)$, the percentile $h^p$ ($0 \leq p \leq 100$), is the $h$ value of a variable $v \in V$, such that at least $p\%$ of the variables $v' \in V$ have $h(v')$ larger than or equal to $h^p$.*

**Definition 10  (Top $p$ Variable Set).** *Given a SAT instance $(V, C)$ and a percentile $h^p$, we call $V(h^p)$ the top $p$ variable set, denoted as $V^p$.*

We use $V_o$ and $V_\delta$ to denote the action variables and transition variables in $V$, respectively. We also define $V_o^p = V_o \cap V^p$, and similarly $V_\delta^p = V_\delta \cap V^p$. Table 1 compares the $h$ values of transition variables and action variables in SAT instances. This table has two parts. For the first part, we list the average and standard deviation of $h$ values for both transition variables and action variables. The data are collected from the first satisfiable SAT instance of the largest solvable planning task in every domain that we consider. From the average $h$ value, it is evident that in most domains transition variables occur more frequently than action variables. Furthermore, the standard deviation of transition variables are in general not only larger than action variables' standard deviation, but also even larger than the expected value of transition variables. The high frequencies of transition variables, along with large standard deviations, are preferred by the VSIDS heuristic and can aid SAT solving, as we discussed earlier.

The second part lists the average $h$ values for transition variables and action variables, in the top $p$ variable set with different values of $p$: 1%, 2%, 5% and 10%. The difference between $V_o^p$ and $V_\delta^p$ is very large. In most domains, transition variables dominate the top variable sets, while action variables exist in the top 10% variable set in only a few domains. One exception is the Airport domain. However, even in this domain, although the average $h$ value of all transition variables is smaller than the average $h$ value of action variables, among the top 1% variables, the average $h$ value of transition variables is larger than the average $h$ value of action variables. Since VSIDS picks the variable with the highest heuristic value, transition variables have higher chances be picked as the decision variables.

### 5.3  Branching Frequency of Transition Variables

In Section 5.2, we considered the difference between transition variables and action variables, in terms of the $h$ values. As mentioned earlier, however, the VSIDS heuristic periodically updates heuristic values of all variables. The dynamic updating of heuristic values is not captured by the above analysis. In the following, we present more direct empirical evidence to show that transition variables are indeed chosen more frequently than action variables for branching, especially at early stages of SAT solving. That is, a SAT solver spends most of its time deciding an appropriate transition plan. This analysis takes into consideration VSIDS's dynamic updating strategy.

We empirically test the probabilities that transition variables and action variables are chosen as branching variables. We measure for every $k$ consecutive decision variables, the number of transition variables ($M_\delta$) and action variables ($M_o$) selected as the decision variables. If all variables are selected equally likely, we should have

$$E(M_\delta) = k \frac{|V_\delta|}{|V_\delta| + |V_o|} \text{ and } E(M_o) = k \frac{|V_o|}{|V_\delta| + |V_o|}, \tag{3}$$

which implies:

$$\frac{E(M_\delta)}{k|V_\delta|} = \frac{E(M_o)}{k|V_o|} \tag{4}$$

| Instances | N | $V_\delta$ | | $V_o$ | | $\overline{h}$ of $V_\delta^p$ | | | | $\overline{h}$ of $V_o^p$ | | | |
|---|---|---|---|---|---|---|---|---|---|---|---|---|---|
| | | $\overline{h}$ | $\sigma$ | $\overline{h}$ | $\sigma$ | 1% | 2% | 5 % | 10% | 1% | 2% | 5 % | 10% |
| Airport-48 | 68 | 8.6 | 7.6 | 19.6 | 12.9 | 98.5 | 75.5 | 59.2 | 23.4 | 39.6 | 35.9 | 34.7 | 32.4 |
| Depot-14 | 12 | 10.5 | 6.0 | 6.3 | 3.3 | 32.5 | 28.6 | 23.7 | 20.9 | - | - | - | - |
| Driverlog-16 | 18 | 32.3 | 11.1 | 5.6 | 3.1 | 43.9 | 34.6 | 26.5 | 23.4 | - | - | - | - |
| Elevator-16 | 15 | 22.2 | 7.4 | 10.2 | 3.8 | 27.0 | 18.9 | 18.9 | 18.9 | - | - | - | - |
| Freecell-6 | 16 | 42.6 | 58.0 | 33.2 | 7.0 | 115.6 | 86.0 | 49.0 | 34.7 | - | - | - | - |
| Openstacks-2 | 23 | 14.1 | 5.2 | 11.5 | 4.3 | 17.2 | 17.2 | 16.2 | 15.2 | - | - | - | - |
| Parcprinter-20 | 19 | 12.0 | 11.8 | 15.5 | 5.9 | 44.3 | 42.8 | 26.7 | 17.8 | 30.0 | 30.0 | 30.0 | 30.0 |
| Pathways-17 | 21 | 5.5 | 8.5 | 12.9 | 3.6 | 33.6 | 26.9 | 15.9 | 14.5 | - | - | - | - |
| Pegsol-25 | 25 | 23.0 | 15.5 | 15.2 | 6.3 | 30.0 | 29.8 | 17.2 | 15.5 | - | - | - | - |
| Pipe-notankage-49 | 12 | 22.9 | 47.1 | 41.3 | 3.4 | 77.1 | 57.5 | 36.4 | 25.3 | - | - | - | - |
| Pipe-tankage-26 | 18 | 58.1 | 116.7 | 50.7 | 12.8 | 266.4 | 174.0 | 86.8 | 56.0 | - | - | - | - |
| Rovers-18 | 12 | 15.5 | 14.6 | 16.0 | 6.9 | 175.1 | 86.0 | 35.5 | 35.5 | - | - | - | - |
| Satellite-13 | 13 | 31.8 | 7.8 | 2.0 | 0.3 | 35.0 | 35.0 | 35.0 | 35.0 | - | - | - | - |
| Scanalyzer-28 | 5 | 113.0 | 151.4 | 8.6 | 1.1 | 242.8 | 175.8 | 129.7 | 129.7 | - | - | - | - |
| Sokoban-6 | 35 | 15.6 | 4.8 | 20.0 | 4.7 | 16.6 | 14.1 | 12.8 | 11.2 | - | - | - | 10.0 |
| Storage-13 | 18 | 4.7 | 1.9 | 6.3 | 1.6 | 10.4 | 10.4 | 9.0 | 8.1 | - | - | - | 9.0 |
| TPP-30 | 11 | 12.3 | 16.1 | 4.8 | 0.7 | 84.2 | 57.8 | 34.4 | 24.6 | - | - | - | - |
| Transport-17 | 22 | 22.8 | 19.0 | 4.5 | 1.1 | 99.6 | 58.1 | 52.0 | 41.6 | - | - | - | - |
| Trucks-13 | 24 | 5.1 | 7.6 | 6.4 | 1.2 | 56.7 | 38.2 | 20.8 | 16.3 | - | - | - | - |
| Woodworking-30 | 4 | 6.2 | 5.1 | 10.2 | 3.5 | 23.1 | 22.1 | 18.9 | 17.2 | - | - | - | 13.1 |
| Zenotravel-16 | 7 | 20.2 | 25.0 | 3.9 | 0.3 | 51.3 | 51.3 | 36.2 | 28.5 | - | - | - | - |

Table 1: The $h$ values of transition variables versus action variables in all domains. Column 'N' is the optimal makespan. Column '$\overline{h}$' is the average and Column '$\sigma$' is the standard deviation. Column '$\overline{h}$ of $V_\delta^p$' and '$\overline{h}$ of $V_o^p$' refer to the average $h$ value of transition variables and action variables in $V^p$, while $p$ equals to 1, 2, 5 or 10. '-' means there is no variable in that percentile range.

We empirically study where we divide the SAT solving process into epochs of length $k = 1000$ each, for all domains from IPC-3 to IPC-6. We present the results of three representative domains in Figure 2, and the results on all domains in Figures 9 and 10 in Appendix B. In each domain, we choose an instance with at least 100,000 decisions. In some domains (e.g. Woodworking), even the biggest instance has only thousands of decisions. In such a case, we choose the instance with the largest number of decisions. For every epoch, we plot the **branching frequency**, which is $\frac{M_\delta}{k|V_\delta|}$ for transition variables and $\frac{M_o}{k|V_o|}$ for action variables, respectively. According to (4), these two branching frequencies should be about the same if the two classes of variables are chosen equally likely.

The results from Openstacks and Zenotravel show clear distinctions between transition variables and action variables. While they are evidently different, the branching frequencies in Openstack has a higher variance. The results of Storage domain show a completely different pattern, where variables do not distinguish by their branching frequencies.

From Figures 9 and 10, it is evident that, for all these instances except Storage-12 and Woodworking-20, the branching frequencies of transition variables are higher than that of action variables. In fact, in many cases, the branching frequencies of transition variables can be up to 10 times higher than those of action variables. In Transport-26 and Zenotravel-15, the difference is orders of magnitude larger. Hence, this empirical study shows that the SAT solvers branch much more frequently on the newly introduced transition variables than the action variables.

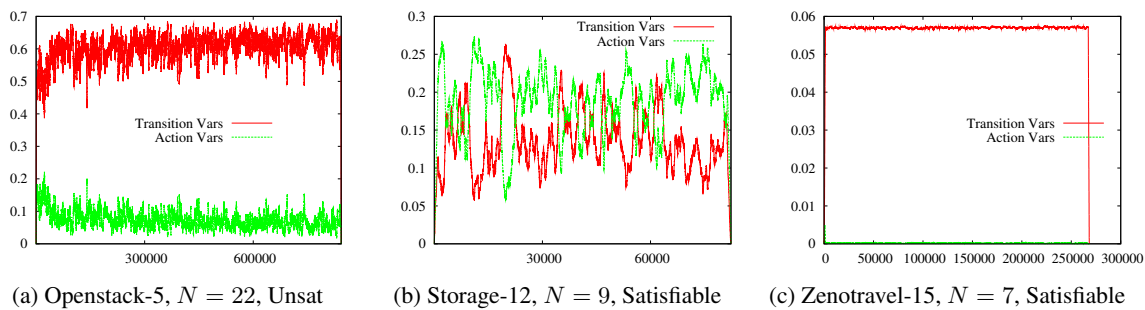

(a) Openstack-5, $N = 22$, Unsat     (b) Storage-12, $N = 9$, Satisfiable     (c) Zenotravel-15, $N = 7$, Satisfiable

Figure 2: Comparison of variable branching frequency (with $k = 1000$) for transition and action variables in solving certain SAT instances on instances from three representative domains: Openstack, Storage and Zenotravel.

## 5.4 Transition Index and SAT Solving Speedup

The behavior of transition variables, as presented above, suggests that there is a correlation between the significance of transition variables and the speedup that SASE achieves. Nevertheless, the study on branching frequency only profiles the connection by showing what happens during SAT solving. Another interesting study should reveal what leads to the speedup in a more direct way. To quantify the analysis, we introduce *transition index*.

As mentioned earlier, the $h$ value does not exactly reflect how VSIDS works, as it updates dynamically throughout SAT solving. Nevertheless, by putting together all variables and studying their $h$ values, the statistics on the population leads to the following definition of the transition index.

**Definition 11 (Transition Index).** *Given a planning problem's SAT instance $(V, C)$, we measure the top $p(0 \le p \le 100)$ variable set, and calculate the transition index of $p$ as follows:*

$$\frac{|V_\delta^p|/|V^p|}{|V_\delta|/|V|}$$

Essentially, the transition index measures the relative density of transition variables in the top variable set. If the distribution of the transition variables is homogeneous under the total ordering based on $h$, $|V_\delta^p|/|V^p|$ should equal to $|V_\delta|/|V|$ for any given $p$. A transition index larger than 1 indicates that the transition variables have a higher-than-normal density in the top $p\%$ variable set. The larger the transition index is, the more often the transition variables occurring in the top $p\%$ variable set.

Given a planning problem's SAT instance, there is correlation between its transition index and the speedup SASE provides. In Figures 3 and 4 we measure such correlation for all the domains from IPC-3 to IPC-6. Each dot in one of the figures refers to an individual planning instance. The y-axis is the speedup of SASE over SatPlan06. The x-axis is the transition index under a given $p$. Bootstrap aggregating (Breiman, 1996) is used for the regression lines. For each measurement, we calculate Spearman's rank correlation coefficient (Myers & Well, 2003), which assesses how well the relationship between two variables can be described using a monotonic function. If there are no repeated data values, a perfect Spearman's correlation coefficient of 1 occurs when each of the variables is a perfect monotone function of the other.

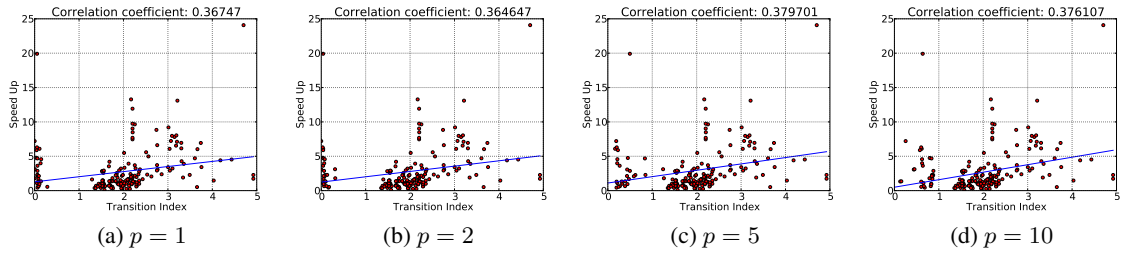

Figure 3: The correlation between SAT solving speedup and the transition index with different $p$. All problem instances are included. We can see a clear cluster of outliers on the bottom-left of each graph, all of which are from the Airport and Rovers domains.

The instances included in Figure 3 are those solved by both SatPlan06 and SASE, with Precosat as the SAT solver. To reduce noise, we do not consider those small instances that both SASE and SatPlan06 spend less than 1 second to solve. In total we have 186 instances. The speedup of each instance is SASE's SAT solving time divided by SatPlan06's SAT solving time, which is greater than 1 in most cases. It can be observed there is a trend that a larger transition index leads to higher speedup. Such a result links the significance of top ranked (high frequency) transition variables to the speedup in SAT solving.

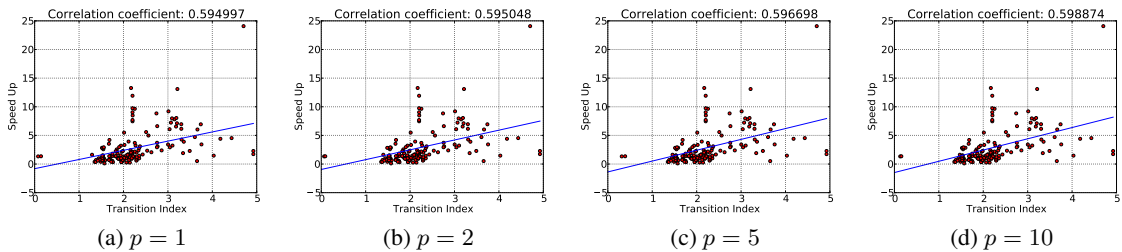

Figure 4: The correlation between SAT solving speedup and the transition index with different $p$. Instances from Airport and Rovers domains are not included.

In Figure 3 there is a cluster of instances of small transition indices (to the bottom-left of each plot in Figure 3). These instances distinguish themselves by having much smaller transition indexes. In fact, it turns out that these instances are all from either the Airport or the Rovers domain and have the same property: there are a very high number of action mutual exclusions, contributing to the majority of clauses. On the other hand, mutual exclusions are binary constraints, which do not contribute significantly to a SAT problem's hardness, because they are trivial for unit propagation. As mentioned earlier, transition index is merely a heuristic to indicate the significance of transition variables. In these instances, the enormous number of action mutual exclusion constraints makes the transition index small. However, they do not make the problems harder, because two-literal clauses are trivial for SAT solving. As a result, we can ignore these outlier instances in our correlation analysis. In Figure 4 we removed those instances from the Airport and Rovers domains, resulting in a total of 159 instances. In this analysis, the correlation becomes even more explicit.

There is, however, one caveat for this study by including all instances from all domains. Some domains have more than thirty instances solved, while in some other domains, we can only solve five. As a result, this study is biased toward those domains with more instances solved. It will be an interesting future study to see how transition index works in a more sophisticated experimental setting, such as by eliminating certain domain specific factors (Hoffmann, Gomes, & Selman, 2006).

| Instances | Before subsumed | | After subsumed | |
|---|---|---|---|---|
| | count | size | count | size |
| Pipesworld-20 | 2548 | 21.72 | 516 | 53.66 |
| Storage-20 | 1449 | 12.46 | 249 | 60.22 |
| Openstack-10 | 221 | 22.44 | 141 | 23.4 |
| Airport-20 | 1024 | 6.45 | 604 | 8.49 |
| Driverlog-15 | 1848 | 2.82 | 1848 | 2.82 |

Table 2: Statistics of action cliques, before and after the subsumed action cliques are reduced. "count" gives the number of action cliques, and "size" is the average size of the action cliques.

## 6. Reducing the Encoding Size of SASE

We now propose several techniques to further reduce the size of SAT instances in SASE. We first represent all mutual exclusions in SASE using a more compact clique representation. We then develop a few new techniques to recognize the special structures of SASE and further reduce the encoding size.

### 6.1 Mutual Exclusion Cliques

Mutual exclusions in SASE naturally define cliques of transitions or actions in which at most one of them can be true at any time step. There are two types of cliques: 1) for each $x \in \mathcal{X}$, $\mathcal{T}(x)$ is a *clique of transitions* enforced by the class E clauses, and 2) for each transition $\delta$ that is not prevailing, $A(\delta)$ is a *clique of actions* enforced by the class H clauses.

It requires $O(n^2)$ clauses to encode all mutexes within a clique of size $n$ in a pair-wise manner. To reduce the number of clauses used, we in SASE use a compact representation (Rintanen, 2006), which uses $\Theta(n \log n)$ auxiliary variables and $\Theta(n \log n)$ clauses. For cliques with large $n$, the reduction in number of clauses will be significant. To show how it works, consider a simple example. Suppose that we have a clique $\{x, y, z\}$ where at most one variable can be true. We introduce auxiliary variables $b_0$ and $b_1$ and clauses $x \Leftrightarrow \overline{b_0} \wedge \overline{b_1}$, $y \Leftrightarrow \overline{b_0} \wedge b_1$ and $z \Leftrightarrow b_0 \wedge \overline{b_1}$.

### 6.2 Reduction Techniques

Action variables form the majority of all variables, and also lead to many clauses to represent action mutual exclusions even if the clique technique is used. Thus, it is important to reduce the number of action variables. We propose three methods when certain structure of a SAS+ planning task is observed.

#### 6.2.1 REDUCING SUBSUMED ACTION CLIQUES

We observe that many action cliques share common elements, while transition cliques do not. In the following, we discuss the case where one action clique is a subset of another. Given two transitions $\delta_1$ and $\delta_2$, if $A(\delta_1) \subseteq A(\delta_2)$, we say clique $A(\delta_1)$ is **subsumed** by clique $A(\delta_2)$.

In preprocessing, for each transition $\delta_1 \in \mathcal{T}$, we check if $A(\delta_1)$ is subsumed by another transition $\delta_2$'s action clique. If so, we do not encode action clique $A(\delta_1)$. In the special case when $A(\delta_1) = A(\delta_2)$ for two transitions $\delta_1$ and $\delta_2$, we only need to encode one of them.

Table 2 presents the number of cliques and their average sizes, before and after reducing action cliques, on some representative problems. The reduction is substantial on most problem domains, except for Driverlog in which no reduction occurred. Note that the average sizes of cliques are increased since smaller ones are subsumed and not encoded.

### 6.2.2 Unary Transition Reduction

Given a transition $\delta$ such that $|\mathcal{T}(\delta)| = 1$, we say that the only action $a$ in $\mathcal{T}(\delta)$ is reducible. Since $a$ is the only action supporting $\delta$, they are logically equivalent. For any such action $a$, we remove $V_{a,t}$ and replace it by $U_{\delta,t}$, for $t = 1, \cdots, N$. An effect of this reduction on a few representative domains can be seen in Table 3.

### 6.2.3 Unary Difference Set Reduction

Besides unary transition variables, an action variable may also be eliminated by two or more transition variables. A frequent pattern is the following: given a transition $\delta$, for all actions in $A(\delta)$, their transition sets differ by only one transition.

**Definition 12** *Given a transition $\delta \in \mathcal{T}$, let $I = \bigcap_{a \in A(\delta)} Trans(a)$. If for every $a \in A(\delta)$, $|Trans(a) \setminus I| = 1$, we call the action set $A(\delta)$ a unary difference set.*

Consider a transition $\delta_1$ with $A(\delta_1) = \{a_1, a_2, \ldots, a_n\}$. If $A(\delta_1)$ is a unary difference set, the transition sets must have the following form:

$$Trans(a_1) = \{\delta_1, \delta_2, \ldots, \delta_k, \theta_1\}$$

$$Trans(a_2) = \{\delta_1, \delta_2, \ldots, \delta_k, \theta_2\}$$

$$\vdots$$

$$Trans(a_n) = \{\delta_1, \delta_2, \ldots, \delta_k, \theta_n\}$$

In this case, we eliminate the action variables for $a_1, \cdots, a_n$ by introducing the following clauses. For each $i, i = 1, \cdots, n$, we replace $V_{a_i,t}$ by $U_{\delta_1,t} \wedge U_{\theta_i,t}$, for $t = 1, \cdots, N$. In such a case, the action variables can be eliminated and represented by only two transition variables. The reason that this reduction can be done is that the $n$ actions are in at least one action clique. The mutual exclusions between these actions maintain the correctness when all but one of the shared transitions are reduced.

Table 3 shows the number of reducible actions in several representative problems. In Zenotravel, all action variables can be eliminated when the two reduction methods are used. In Openstack and Storage, there is only one type of reduction that can be applied.

| Instances | $|\mathcal{O}|$ | $R_1$ | $R_2$ | % |
|-----------|------|------|------|--------|
| Zeno-15 | 9420 | 1800 | 7620 | 100.00 |
| Pathway-15 | 1174 | 173 | 810 | 83.73 |
| Trucks-15 | 3168 | 36 | 300 | 10.61 |
| Openstack-10 | 1660 | 0 | 400 | 24.10 |
| Storage-10 | 846 | 540 | 0 | 63.83 |

Table 3: Number of reducible actions in representative instances. Columns '$R_1$' and '$R_2$' give the number of action variables reduced, by unary transition reduction and unary difference set reduction, respectively. Column '%' is the percentage of the actions reduced by both methods combined.

## 7. Experimental Analysis and Results

We experimentally analyzed the performance of planning using SASE in comparison against many state-of-the-art planners. We tested all problem instances of STRIPS domains in IPC-3 to IPC-6. PSR and Philosophers were not included because they have derived facts, which cannot be handled correctly by any of the planners tested. We used the parser by Fast-Downward (Helmert, 2006, 2008) to generate the SAS+ formalism from STRIPS inputs. The preprocessing and encoding parts of SASE were implemented in Python2.6. All the instances were based on grounded STRIPS. In nearly all cases, the problem solving took much longer time than the pre-processing, thus we only reported the overall running time.

We ran all experiments on a PC workstation with a 2.3 GHz AMD Quad-Core Opteron processor. The running time for each instance was set to 1800 seconds, and the memory was limited to 4GB. For all planners, the running time included parsing, preprocessing and problem solving. The memory consumption was the peak memory usage reported by the SAT solvers.

### 7.1 Comparison Results

Precosat (build236) (Biere, 2009), the winner of the application track in the SAT'09 competition, was used as the SAT solver for most planners that we tested and compared. Besides Precosat, we also used CryptoMinisat (Soos, Nohl, & Castelluccia, 2009), the winner of SAT Race 2010, as the underlying solver of SatPlan06 and SASE. The nine planners considered are listed as follows.

1. **SP06** and **SP06-Crypto**. They are the original SatPlan06 planner (Kautz et al., 2006), only with the underlying SAT solver changed to Precosat and CryptoMinisat, respectively.

2. **SASE** and **SASE-Crypto**. They are SASE encoding introduced in this paper, with all the optimization methods turned on. The underlying SAT solvers are Precosat and CryptoMinisat, respectively.

3. **SP06L**. It is SatPlan06 (Kautz et al., 2006) with long-distance mutual exclusion (londex) (Chen et al., 2009). We compared against londex since it also derives transition information from the SAS+ formalism. We used domain transition graphs from Fast-Downward's parser to derive londex information.

4. **SP06C**. It is SatPlan06 with the clique technique (Rintanen, 2006) to represent the mutual exclusions. The clique information was obtained via Fast-Downward. Note that due to the different grounding strategies by SatPlan06 and Fast-Downward, not all of the mutual exclusions defined in SatPlan06 could be covered by cliques.

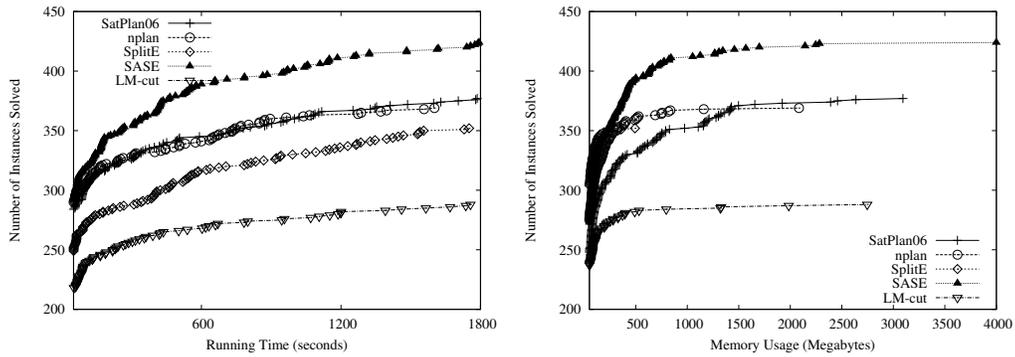

Figure 5: The results on different planners. We include the default version of every planner. The figures show the number of problems solved by each planner, with increasing limits on running time or memory consumption.

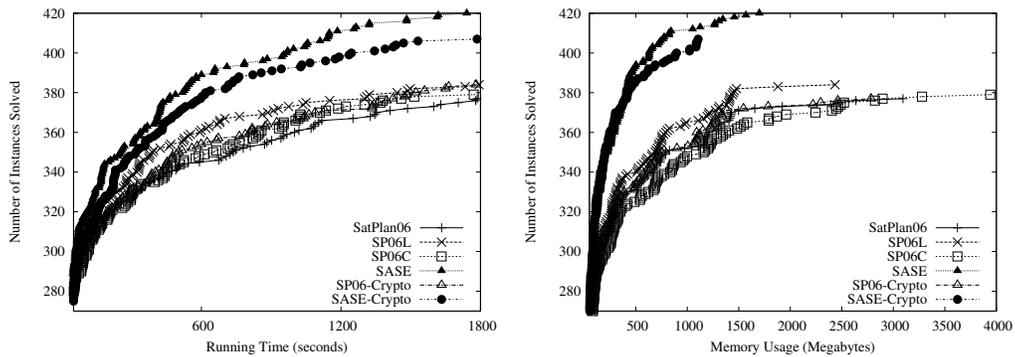

Figure 6: The results on the variants of SatPlan06 and SASE. The data is presented as the number of problems solved by each planner, with increasing limits on running time or memory consumption.

5. **nplan**. The nplan solver (Rintanen et al., 2006) is set to use $\forall$-step to generate plans with the same optimality metric as other planners. The executable is the most recent release from nplan's homepage. The build-in SAT solver is changed to Precosat.

6. **SplitE**. It is the split encoding (Robinson et al., 2009) using Precosat. We have obtained source code from the authors and recompiled it on our 64bit Linux workstation.

7. **LM-cut**. This is a sequential optimal planner, using LM-Cut heuristic (Helmert & Domshlak, 2009) and A* search. We used the implementation in Fast-Downward.

We present the results as two sets of planners, as in Figures 5 and 6, respectively. For both sets of data, we show the number of instances that are solvable in the testing domains, with respect to the given time limit and memory limit.

Figure 5 compares the results of several different solvers using their original version. Our data suggests that SASE has clear advantages. LM-cut is the least efficient, although the comparison is not very meaningful as it uses an optimization metric different from other planners. For running time and memory consumption, SASE is clearly superior to all the other planners. Among all planners, nplan is slightly better than the others on smaller instances, but on larger instances, SatPlan06 becomes more competitive.

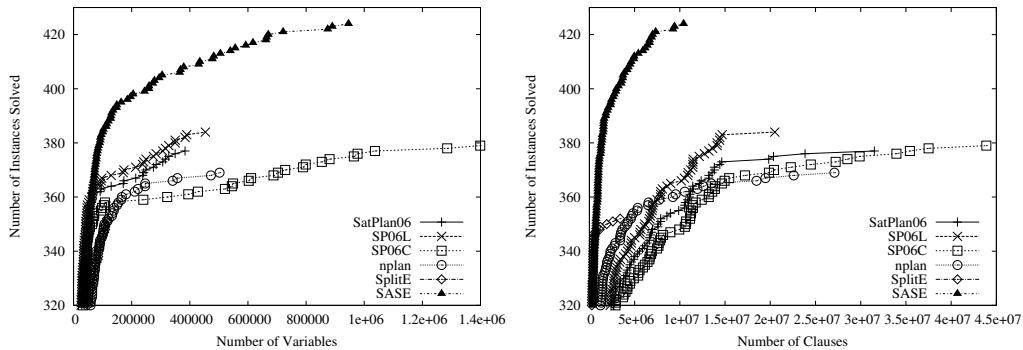

Figure 7: Number of problems solved by each planner, with increasing limits on number of variables and number of clauses.

In Figure 6, we compare the results of different variants of both SatPlan06 and SASE. Both SP06L and SP06C extend SatPlan06 with additional techniques. They in general make little improvements over the original SatPlan06.

For SAT based planners, we present in Figure 7 the number of instances that are solvable with increasing limits on the number of variables and number of clauses. Note that the curves are slightly affected by the given time and memory limit, thus efficient planners like SASE stops at a smaller number of clauses. The results show SASE has an advantage in terms of the number of variables and number of clauses over the other planners.

Table 4 presents the number of instances solved in each planning domain, within the given time and memory limit. In general, SASE solved more instances than the other planners. Due to some programming bugs, nplan could not find the correct solutions with the optimal makespan in domains Openstacks, Rovers and Storage. The SplitE parser could not handle problems in Airport and Pathways. Therefore, we did not evaluate the corresponding encoding on those benchmarks. Although LM-Cut overall solved fewer instances, in a few domains it performed better than all the SAT based planners. These domains seemed to allow less concurrencies between action. In particular, domains Openstacks and Sokoban only have plans that are *strictly* sequential, meaning that there are no actions that can be executed at the same time step. The plans for such instances often require more time steps, making them more challenging for SAT-based planners.

Both SP06L and SP06C used Fast-Downward's parser to obtain domain transition graph information. Therefore, for SP06C and SP06L, it took too much time to pre-process grounded STRIPS instances twice (one by Fast-Downward and one by original SP06). In consequence, the efficiency of londex or clique representation may not compensate for pre-processing time, leading to slightly worse performance than the original SP06 in a few instances. For example, londex was helpful in TPP, but not in Trucks and Scanalyzer. The clique representation was very helpful in Airport domain, with 10 more instances solved, but did not help much in Pegsol and Satellite.

| Domain | SP06 | SP06L | SP06C | nplan | SplitE | SASE | SP06$^c$ | SASE$^c$ | SASE$^0$ | LM |
|---|---|---|---|---|---|---|---|---|---|---|
| Airport | 35 | 38 | 39 | 20 | 0 | **46** | 38 | 42 | 39 | 27 |
| Depot | 17 | 16 | 16 | **19** | 17 | 17 | 17 | 15 | 14 | 7 |
| Driverlog | 16 | 16 | 16 | **17** | **17** | **17** | **17** | **17** | 16 | 13 |
| Elevator | **30** | **30** | **30** | **30** | **30** | **30** | **30** | **30** | **30** | 19 |
| Freecell | 5 | 4 | 5 | **6** | 5 | **6** | 4 | **6** | 6 | 5 |
| Openstacks | 5 | 5 | 5 | 0 | 5 | 5 | 5 | 5 | 5 | **20** |
| Parcprinter | 29 | 29 | 29 | **30** | 29 | **30** | 29 | **30** | 30 | 21 |
| Pathways | 11 | 11 | 11 | **12** | 0 | **12** | 9 | 10 | 12 | 5 |
| Pegsol | 21 | 21 | 21 | 21 | 22 | 24 | 18 | 19 | 22 | **27** |
| Pipe-notankage | 38 | 37 | 31 | **40** | 37 | 37 | 38 | 35 | 38 | 17 |
| Pipe-tankage | 16 | 16 | 16 | 22 | 10 | **26** | 13 | 23 | 16 | 11 |
| Rovers | 13 | 13 | 13 | 0 | **18** | 14 | 13 | 17 | 16 | 7 |
| Satellite | 17 | 17 | 17 | **18** | 16 | **18** | 17 | 17 | 15 | 7 |
| Scanalyzer | 15 | 14 | 14 | **18** | 13 | **18** | 16 | 17 | 17 | 7 |
| Sokoban | 5 | 5 | 3 | 11 | 5 | 5 | 5 | 5 | 5 | **24** |
| Storage | 15 | 15 | 15 | 0 | **16** | 15 | 15 | 15 | 15 | 15 |
| TPP | 27 | **30** | 29 | 28 | 25 | **30** | 28 | 29 | 29 | 6 |
| Transport | 19 | 16 | 19 | **22** | 18 | **22** | 19 | 21 | 20 | 12 |
| Trucks | 7 | 6 | 5 | **10** | 8 | 8 | 7 | 8 | 7 | **10** |
| Woodworking | **30** | **30** | **30** | **30** | **30** | **30** | **30** | **30** | **30** | 16 |
| Zenotravel | 15 | 15 | 15 | 15 | 15 | **16** | **16** | **16** | **16** | 12 |
| Total | 386 | 384 | 379 | 369 | 336 | **426** | 384 | 407 | 398 | 288 |

Table 4: Number of instances solved in each domain within 1800 seconds. SP06$^c$, SASE$^c$ and LM are short for SP06-Crypto, SASE-Crypto and LM-Cut. Column 'SASE$^0$' is the result of SASE without any reduction and optimization.

Comparing with nplan, in general SASE was better, but nplan performed better than SASE on those domains with few concurrencies. For example, both Sokoban and Trucks have only one action at nearly every time step. We believe the reason should be the way nplan encodes all mutual exclusions as linear encoding (Rintanen et al., 2006), which could be further used to improve SASE.

SplitE in general was slightly worse than SP06. It won over SP06 on 5 domains and SP06 was superior to SplitE on 6 domains. Overall, SplitE was competitive not to nplan or SASE. Rovers however was the only domain where SplitE performed better than all others. Although CryptoMinisat performed better than Precosat in SAT Race 2010, it was not as good for planning problems. For both SASE and SP06, CryptoMinisat solved fewer instances.

## 7.2 Ablation Study

Figure 8 shows the number of solvable problems from all the problems in IPC-3 to IPC-6, with increasing limits on running time, memory consumption, number of variables and number of clauses. Precosat was used for all planners. Running time was the total time including preprocessing and problem solving. Memory usage was based on the status report by Precosat. Under the maximum CPU time (1800s) and memory limit (4Gb), when both clique representation and reduction techniques were not used, SASE solved 398 instances. By turning on either the clique representation or the action reduction technique, SASE solved 416 and 405 instances, respectively. When both clique and action reduction techniques were turned on, SASE solves 426 instances.

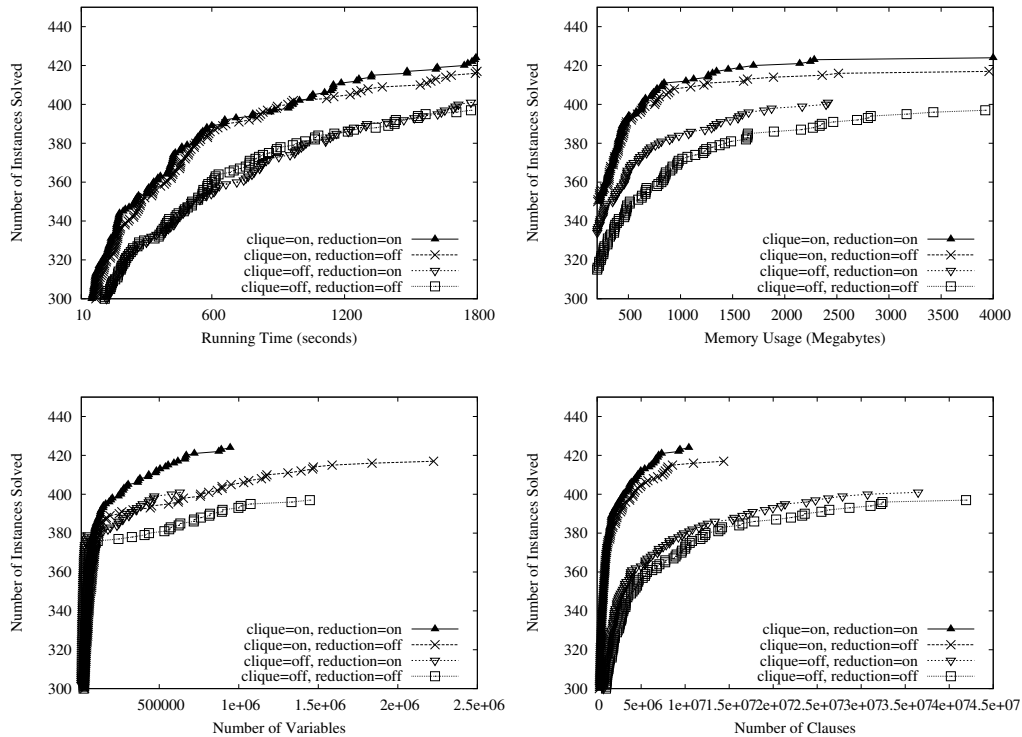

Figure 8: The results of SASE with the clique and reduction methods turned on or off.

The reduction method improved upon problem solving time, as well as the clique representation. The clique representation provided a substantial improvement to memory consumption, followed by action reduction. For numbers of clauses, the clique technique gave a significant reduction. Finally, both techniques helped reduce the number of variables.

## 8. Conclusions and Future Research

In this paper, we developed a novel SAS+ based SAT encoding scheme SASE, and showed that it improves the efficiency of STRIPS based SAT encodings in terms of both time and memory. When compared with the state-of-the-art SAT based planners, SASE has clear advantages as shown by our experimental analysis. We proved the correctness of the SASE encoding by showing that there is an isomorphism between the solution plans of SASE and the solution plans of SatPlan06. We further analyzed the search space structure of SASE, and explained why it is more efficient. Below, we briefly discuss some related work and highlight several directions for future research.

### 8.1 Other Semantics and Encodings

Many enhancements have been developed for SAT based planning since it was introduced (Kautz & Selman, 1992). The split action representation (Kautz & Selman, 1992; Ernst et al., 1997) uses a conjunction of multiple variables to represent an action. The optimality is, however, lost. Robinson et al. (2009) propose a new way of doing splitting without sacrificing the optimality. The results show that this method has advantages over SatPlan06 (Kautz et al., 2006). There are many published works with thorough analysis, or improvements along this line of research on SatPlan family encodings. In particular, the power of mutual exclusion, in the context of planning as SAT, has attracted interests (Chen et al., 2009; Sideris & Dimopoulos, 2010). A new encoding scheme called SMP is proposed, which preserves the merits of londex and shows certain advantages over the existing encoding schemes (Sideris & Dimopoulos, 2010).

The planner family of SatPlan and its variants are step-optimal. The step-optimality semantics, along with a relaxed parallel semantics, are formalized as ∀-step and ∃-step, respectively (Dimopoulos, Nebel, & Koehler, 1997; Rintanen et al., 2006). ∃-step enforces weaker mutual exclusions than ∀-step, thus may lead to reduced running time due to fewer calls to the SAT solver. As a trade-off, it loses the optimality of time steps. The semantics in both SatPlan06 and SASE are ∀-step. The research on various kinds of semantics are orthogonal to our contribution in SASE, and the idea of SASE can be migrated to new semantics.

Since SAT based planning has also been applied to sequential planning, the idea of SASE can also be extended to this field. The first planner of this kind is MEDIC (Ernst et al., 1997), which extends the idea of splitted action representation. This study shows that for sequential planning, splitting yields very competitive planners. It has also been proposed to utilize the advantages of sequential and parallel planning (Büttner & Rintanen, 2005).

## 8.2 Additional Techniques for Planning as SAT

The tremendous amount of two-literal clauses (such as the mutual exclusion clauses in the case of planning) is a key challenge to approaches based on satisfiability tests. It has been proposed to mitigate the burden of encoding them by recognizing certain structures (Brafman, 2001). In traditional SAT planning systems like SatPlan06, the mutual exclusions are encoded in a quadratic manner. Rintanen proposes a log size technique (Rintanen, 2006), called clique representation, for the mutual exclusion constraints, and later a linear size one (Rintanen et al., 2006). The mutual exclusions in SASE are represented by the clique representation. The log size of the clique representation is supposed to be less compact than the linear encoding. Our results, however, have shown that SASE is in general more compact. This is mainly due to the compactness of SAS+ formalism. It is certainly an open question whether the linear size encoding technique can be adopted to further improve SASE.

There are also techniques beyond encoding to boost SAT-based planning. Rintanen introduces how to incorporate symmetry information into SAT instances (Rintanen, 2003). MaxPlan (Xing, Chen, & Zhang, 2006) does a planning graph analysis to find an upper bound on the optimal make span and then does SAT queries using decreasing time steps, until it meets an unsatisfiable SAT instance. A lemma reusing method is proposed (Nabeshima, Soh, Inoue, & Iwanuma, 2006) to reuse the learnt clauses across multiple SAT solvings. A multiple-step query strategy is introduced (Ray & Ginsberg, 2008), which however asks for modified SAT procedures. All these boosting methods are proposed in the context of STRIPS based planning. It is interesting to incorporate these techniques into SASE and study if they can further improve the performance.

## 8.3 More Understanding of Structure in General SAT Instances

SAT is intrinsically hard. The performance of modern SAT solvers improves constantly. It is therefore interesting and important to understand why SAT solvers work well on certain instances, and furthermore, what makes a SAT instance easy or hard. There is much prior research that tries to obtain such an understanding, including backdoor set (Williams, Gomes, & Selman, 2003) and backbone (Monasson, Zecchina, Kirkpatrick, Selman, & Troyansky, 1999; Zhang, Rangan, & Looks, 2003; Zhang, 2004). Backdoor set variables are a set of variables, such that when these variables are assigned, other variables' assignments can be derived in polynomial time. Backbone variables are those variables that have the same assignment in all valid solutions, which can be further exploited to improve SAT solving efficiency. A recent study in the context of planning reveals that there is clear correlations between the SAT solving efficiency and goal asymmetry (Hoffmann et al., 2006).

It will be interesting to see if there are connections between SASE's problem structure and those theories above. For example, is the improvement of SASE because it can lead to a smaller backdoor set? Second, we have shown that the efficiency of SASE is a result of transition variables' significance, and there is strong correlation between the speedup from SASE and the transition index. It is interesting to investigate if similar variable set and predictive index can be automatically found for general SAT solving.

Finally, given the efficiency of SASE, it is promising to apply it to other SAT-based planning approaches, such as those for complex planning with preferences (Giunchiglia & Maratea, 2007) and temporal features (Huang et al., 2009).

## 9. Acknowledgments

This research was supported by National Science Foundation of the United States under grants NeTS-1017701, DBI-0743797, IIS-0713109, and a Microsoft Research New Faculty Fellowship. We thank Joerg Hoffmann, Jussi Rintanen and several anonymous reviewers for the helpful comments. We particularly thank Malte Helmert for making the SAS+ parser available. We also thank the computing resource and supports from engineering IT group of Washington University in St. Louis.

## Appendix A. Proofs

**Lemma 1** Given a STRIPS task $\Psi = (\mathcal{F}, \mathcal{A}, \varphi_{\mathcal{I}}, \varphi_{\mathcal{G}})$, a time step $N$, and its PE SAT instance $PE(\Psi, N) = (V, C)$, suppose there is a satisfiable solution denoted as $\Gamma$, a fact $f \in \mathcal{F}$, and $t \in [1, N]$ such that: 1) $\Gamma(W_{dum_f,t}) = \bot$, 2) $\Gamma(W_{f,t}) = \top$, and 3) $\forall a \in \text{DEL}(f)$, $\Gamma(W_{a,t}) = \bot$, then we can construct an alternative solution $\Gamma'$ to $PE(\Psi, N)$ as follows:

$$\Gamma'(v) = \begin{cases} \top, \, v = W_{dum_f,t}, v \in V \\ \Gamma(v), \, v \neq W_{dum_f,t}, v \in V \end{cases} \tag{5}$$

**Proof** We show that $\Gamma'$ satisfies every clause in $C$ just as $\Gamma$ does. Since all variables but $W_{dum_f,t}$ keep the same value, we only need to examine those clauses that have $W_{dum_f,t}$ in them. According to the definition of PE, $W_{dum_f,t}$ may exist in three types of clauses:

1. Clauses for add effects. In this case, the clauses are of the form $W_{f,t+1} \rightarrow (W_{dum_f,t} \vee W_{a_1,t} \vee \cdots \vee W_{a_m,t})$, which is equivalent to $\overline{W}_{f,t+1} \vee W_{dum_f,t} \vee W_{a_1,t} \vee \cdots \vee W_{a_m,t}$. Since $\Gamma'(W_{dum_f,t}) = \top$, such clauses are still true.

2. Clauses for preconditions. In this case, the clauses are of the form $W_{dum_f,t} \rightarrow W_{f,t}$, which is equivalent to $\overline{W_{dum_f,t}} \vee W_{f,t}$. Since $\Gamma'(W_{f,t}) = \top$, these clauses remain true for $\Gamma'$.

3. Clauses of mutual exclusion between actions. Without loss of generality, let us denote such a clause $\overline{W_{dum_f,t}} \vee \overline{W_{a,t}}$. For a given $f$, the actions in all such clauses are mutex with $dum_f$, because $f$ is their delete effect. According to the construction, since $\Gamma'(W_{a,t}) = \Gamma(W_{a,t}) = \bot$, all such clauses are true.

The three cases above conclude that all clauses that include $W_{dum_f,t}$ are satisfied by $\Gamma'$. Therefore, $\Gamma'$ is also a solution to PE. $\qquad\square$

**Lemma 2** Given a STRIPS task $\Psi = (\mathcal{F}, \mathcal{A}, \varphi_{\mathcal{I}}, \varphi_{\mathcal{G}})$, a time step $N$, and its PE SAT instance $\text{PE}(\Psi, N) = (V, C)$, suppose there is a satisfiable solution denoted as $\Gamma$, a fact $f \in \mathcal{F}$, and $t \in [1, N]$ such that: 1) $\Gamma(W_{f,t}) = \bot$, 2) there exists an action $a \in \text{ADD}(f)$ such that $\Gamma(W_{a,t-1}) = \top$, then we can construct an alternative solution $\Gamma'$ to $\text{PE}(\Psi, N)$ as follows:

$$\Gamma'(v) = \begin{cases} \top, v = W_{f,t}, v \in V \\ \Gamma(v), v \neq W_{f,t}, v \in V \end{cases} \tag{6}$$

**Proof** We show that $\Gamma'$ makes each clause in $C$ to be true. Since all variables but $W_{f,t}$ keep the same value, we only need to look at those clauses that have $W_{f,t}$ in them. According to the definition of PE, $W_{f,t}$ may exist in three types of clauses.

1. Clauses for add effects. In this case, $f$ is an add effect of multiple actions. Let us write this clauses as $W_{f,t} \rightarrow (W_{a_1,t-1} \vee W_{a_2,t-1} \vee \cdots \vee W_{a_m,t-1})$, which is $\overline{W_{f,t}} \vee W_{a_1,t-1} \vee W_{a_2,t-1} \vee \cdots \vee W_{a_m,t-1}$. Since there exists an action $a \in \text{ADD}(f)$ such that $\Gamma(W_{a,t-1}) = \top$, the clause is still true in $\Gamma'$.

2. Clauses for preconditions. In this case, $f$ is a precondition of an action $b$. This clause is written as $W_{b,t} \rightarrow W_{f,t}$, which is equivalent to $\overline{W_{b,t}} \vee W_{f,t}$. Since $\Gamma'(W_{f,t}) = \top$, this clause is still true.

3. Clauses of fact mutex. Without loss of generality, consider a fact $g$ that is mutex with $f$. The corresponding clause will be $\overline{W_{f,t}} \vee \overline{W_{g,t}}$. Since $\Gamma'(W_{f,t}) = \top$, this clause is true if $\Gamma'(W_{g,t}) = \bot$.

   We now suppose $\Gamma'(W_{g,t}) = \top$ and show that it leads to a contradiction. According to clauses of class III, there must be a variable $W_{b,t-1}$, such that $g \in add(b)$ and $\Gamma'(W_{b,t-1}) = \top$. According to the definition of mutex, two facts are mutex only when every pair of the actions that add them are mutex. Thus, $W_{a,t-1}$ and $W_{b,t-1}$ are mutex. Therefore, $\Gamma'(W_{a,t-1}) = \top$ and $\Gamma'(W_{b,t-1}) = \top$, leading to a contradiction. As a result, $\Gamma'(W_{g,t}) = \bot$, and consequently this clause is satisfied.

The three cases above conclude that all clauses that include $W_{f,t}$ are satisfied by $\Gamma'$. Therefore, $\Gamma'$ is also a solution to PE. $\qquad\square$

**Lemma 3** *Given a SAS+ planning task $\Pi = (\mathcal{X}, \mathcal{O}, s_{\mathcal{I}}, s_{\mathcal{G}})$ and its equivalent STRIPS task $\Psi = (\mathcal{F}, \mathcal{A}, \varphi_{\mathcal{I}}, \varphi_{\mathcal{G}})$, suppose we have actions $a, b \in \mathcal{O}$, and their equivalent actions $a', b' \in \mathcal{A}$ (i.e. $a = \phi_a(a')$ and $b = \phi_a(b')$), $a$ and $b$ are S-mutex iff $a'$ and $b'$ are P-mutex.*

**Proof**  We construct the proof by studying it on both directions. Based on Proposition 1, we only consider *inconsistent effects* and *interference* mutex in P-mutex.

$\Rightarrow$*: if $a'$ and $b'$ are P-mutex in $\Psi$, $a$ and $b$ are S-mutex in $\Pi$.*

Since $a'$ and $b'$ are P-mutex, one either deletes precondition or add-effect of the other. Without loss of generality, suppose $a'$ deletes $f$ (i.e. $f \in del(a') \cap pre(b')$). Consequently, there must be a transition $\delta_1 = \delta^x_{f \rightarrow h} \in T(a)$ such that $f \neq h$ and $\delta_2 = \delta^x_{f \rightarrow g} \in T(b)$. There are two cases to be considered.

1) $\delta_1 \neq \delta_2$. $\delta_1$ and $\delta_2$ are mutex transitions by Definition 3, since they both change their value from $f$. Therefore, $a$ and $b$ are S-mutex, according to the second condition in Definition 5.

2) $\delta_1 = \delta_2$. In this case, $a$ and $b$ are S-mutex by the first condition of Definition 5.

Based on the two cases, we conclude that $a$ and $b$ are S-mutex. A similar argument applies to the case when one action deletes the other's add-effect.

$\Leftarrow$*: if $a$ and $b$ are S-mutex in $\Pi$, $a'$ and $b'$ are P-mutex in $\Psi$.*

If two actions $a$ and $b$ are S-mutex in $\Pi$, there are two cases.

1) There exists a transition $\delta$, which is in both $T(a)$ and $T(b)$. Consequently, $a'$ and $b'$ deletes each other's precondition and thus they are P-mutex.

2) There exist two distinct transitions $\delta_1 \in T(a), \delta_2 \in T(b)$ and a multi-valued variable $x \in \mathcal{X}$, such that $\{\delta_1, \delta_2\} \subseteq T(x)$. Let us denote these two transitions as $\delta^x_{v_1 \rightarrow v_2}$ and $\delta^x_{v_3 \rightarrow v_4}$. In such a case, suppose $\delta^x_{v_1 \rightarrow v_2}$ and $\delta^x_{v_3 \rightarrow v_4}$ are allowed to be executed in parallel in a STRIPS plan. It obviously leads to a contradiction, since $v_1, v_2, v_3, v_4 \in Dom(x)$ are values of the same multi-valued variable, and by the definition of SAS+ formalism, only one of them can be true at the same time. Therefore, the preconditions of $a'$ and $b'$ must be mutex, and hence $a'$ and $b'$ are P-mutex.  □

**Theorem 1** *Given a STRIPS task $\Psi$ and a SAS+ task $\Pi$ that are equivalent, for a time step bound $N$, if $\text{PE}(\Psi, N)$ is satisfiable, $\text{SASE}(\Pi, N)$ is also satisfiable.*

**Proof**  Since $\text{PE}(\Psi, N)$ is satisfiable, we denote one of its solutions as $\Gamma_\Psi$. We first present how to construct an assignment to $\text{SASE}(\Pi, N)$ from $\Gamma_\Psi$. Next, we prove that this constructed assignment satisfies every clause in $\text{SASE}(\Pi, N)$.

**Construction.**  There are two steps for the construction. According to Lemmas 1 and 2, there are in general some free variables in $\Gamma_\Psi$. In the first step, we construct an alternative solution to $\text{PE}(\Psi, N)$ by changing all free variables in $\Gamma_\Psi$ to be true according to Lemmas 1 and 2. Let us denote the resulting solution as $\Gamma'_\Psi$. Then, we construct an assignment for $\text{SASE}(\Pi, N)$ from $\Gamma'_\Psi$. The value of each variable in $\Gamma_\Pi$ is defined as follows.

1. For every $a \in \mathcal{O}$ (which is also in $\mathcal{A}$)[1], we let $U_{a,t} = W_{a,t}$.

2. For every transition $\delta_{f \rightarrow g} \in \mathcal{T}$, if $W_{f,t} = \top$ and $W_{g,t+1} = \top$ in $\Gamma'_\Psi$, we set $U_{x,f,g,t} = \top$ in $\Gamma_\Pi$.

---

[1] For simplicity, we use $a$ to denote the same action in $\mathcal{A}$ instead of using $\phi_a(a)$.

**Satisfiability.** We prove that every individual clause in SASE is satisfied by $\Gamma_\Pi$. There are eight types of clauses.

1. (Forward progression). According to our construction, we need to show that, for any $t \in [1, N-2]$,

$$\forall \delta_{h \to f}^x \in \mathcal{T}, (W_{h,t} \wedge W_{f,t+1}) \to \bigvee_{\forall g, \delta_{f \to g}^x \in \mathcal{T}} (W_{f,t+1} \wedge W_{g,t+2}) \tag{7}$$

If $\Gamma'_\Psi(W_{f,t+1}) = \bot$, then (7) is satisfied by $\Gamma'_\Psi$. If $\Gamma'_\Psi(W_{f,t+1}) = \top$, we consider an action set $Y = \{dum_f\} \cup \text{DEL}(f)$, which is a subset of $Trans(\delta_{f \to g}^x)$. There are two possibilities.

- For every action $a \in Y$, $\Gamma'_\Psi(W_{a,t+1}) = \bot$. In such a case, $W_{dum_f, t+1}$ and $W_{f,t+2}$ are free variables according to Lemmas 1 and 2, respectively. Therefore, according to the construction in $\Gamma'_\Psi$, which assigns all free variables to true, variables $W_{f,t+1}$, $W_{f,t+2}$ and $W_{dum_f, t+1}$ are all $\top$. In addition, $\delta_{f \to f}$ is always in $\mathcal{T}$, meaning $W_{f,t+2}$ is included in the right hand side of (7). Therefore, (7) is satisfied by $\Gamma'_\Psi$.

- There exists an action $a \in Y$, such that $\Gamma'_\Psi(W_{a,t+1}) = \top$. In such a case, let us consider an arbitrary fact $g \in add(a)$. If $\Gamma'_\Psi(W_{g,t+2}) = \top$, then (7) is satisfied by $\Gamma'_\Psi$. Otherwise, according to Lemma 2, $W_{g,t+2}$ is a free variable and $W_{g,t+2}$ is already set to true in our construction of $\Gamma'_\Psi$. Therefore, $\Gamma'_\Psi$ satisfies (7).

2. (Regression). According to our construction, we need to show that, for any $t \in [2, N-1]$,

$$\forall \delta_{f \to g}^x \in \mathcal{T}, (W_{f,t} \wedge W_{g,t+1}) \to \bigvee_{\forall h, \delta_{h \to f}^x \in \mathcal{T}(x)} (W_{h,t-1} \wedge W_{f,t}) \tag{8}$$

Consider clauses of class III (add effect) in PE. These clauses indicate that for each fact $f \in \mathcal{F}$, $W_{f,t}$ implies a disjunction of $W_{a,t-1}$ for all actions $a$ such that $f \in add(a)$. Thus, for a given $f$, the following clauses are included in PE, which are satisfied by $\Gamma'_\Psi$:

$$W_{f,t} \to \bigvee_{\forall a \in \text{ADD}(f)} W_{a,t-1}. \tag{9}$$

For a given $f$, we consider the action set $\bigcup_{\forall h} A(\delta_{h \to f})$, denoted as $Z$. Since $\text{ADD}(f) \subseteq Z$,

$$W_{f,t} \to \bigvee_{\forall a \in Z} W_{a,t-1} \tag{10}$$

For any transition $\delta_{h \to f}^x$, for each action $a \in A(\delta_{h \to f}^x)$, since $h \in pre(a)$, $\Gamma'_\Psi$ satisfies $W_{a,t-1} \to W_{h,t-1}$. Therefore, for each $h \in pre(a)$, we have

$$\bigvee_{\forall a \in A(\delta_{h \to f}^x)} W_{a,t-1} \to W_{h,t-1}. \tag{11}$$

By expanding set $Z$, we can convert (10) to:

$$W_{f,t} \to \bigvee_{\forall h, \delta^x_{h \to f} \in \mathcal{T}} (\bigvee_{\forall a \in A(\delta^x_{h \to f})} W_{a,t-1}). \tag{12}$$

By combining (11) and (12), we have:

$$W_{f,t} \to \bigvee_{\forall h, \delta^x_{h \to f} \in \mathcal{T}} W_{h,t-1}, \tag{13}$$

which implies

$$W_{f,t} \to \bigvee_{\forall h, \delta^x_{h \to f} \in \mathcal{T}} (W_{h,t-1} \land W_{f,t}). \tag{14}$$

From (14), we can see that the clauses of regression in (8) are true.

3. (Initial state). We need to show that for each variable $x$ in $\mathcal{X}$ such that $s_{\mathcal{I}}(x) = f$:

$$\bigvee_{\forall g, \delta_{f \to g} \in \mathcal{T}} U_{f,g,1} \tag{15}$$

According to our construction, (15) becomes:

$$\bigvee_{\forall g, \delta_{f \to g} \in \mathcal{T}} (W_{f,1} \land W_{g,2}),$$

which is equivalent to:

$$W_{f,1} \land (\bigvee_{\forall g, \delta_{f \to g} \in T(x)} W_{g,2}) \tag{16}$$

Since $f$ is in the initial state, $\Gamma_\Psi(W_{f,1}) = \Gamma'_\Psi(W_{f,1}) = \top$. Therefore the first part of the conjunction in (16) is true. The rest part of (16) can be seen to be true following a similar argument as that for the progression case.

4. (Goal). The goal clauses can be shown in a similar way as that for the initial state clauses.

5. (Composition of actions). The clauses we want to prove are, for any action $a$, $U_{a,t} \to \bigwedge_{\forall \delta \in Trans(a)} U_{\delta,t}$, or equivalently, $\overline{U_{a,t}} \lor \bigwedge_{\forall \delta \in Trans(a)} U_{\delta,t}$.

Suppose $Trans(a) = \{\delta_{f_1 \to g_1}, \delta_{f_2 \to g_2}, \dots, \delta_{f_m \to g_m}\}$. The clause we need to show becomes:

$$(\overline{W_{a,t}} \lor W_{f_1,t}) \land (\overline{W_{a,t}} \lor W_{g_1,t}) \land (\overline{W_{a,t}} \lor W_{f_2,t}) \land (\overline{W_{a,t}} \lor W_{g_2,t}) \land \dots$$
$$\land (\overline{W_{a,t}} \lor W_{f_m,t}) \land (\overline{W_{a,t}} \lor W_{g_m,t}) \tag{17}$$

Let us call these two-literal disjunctions in (17) as sub-clauses. All those $\overline{W_{a,t}} \lor W_{f_i,t}$ sub-clauses in (17) are exactly the same as the precondition clause (class IV) in PE. So all $\overline{W_{a,t}} \lor W_{f_i,t}$ in (17) are satisfied.

Next, let us consider those $\overline{W_{a,t}} \lor W_{g_i,t}$ sub-clauses. For any $g = g_i, i = 1, \dots, m$. There are four cases where $W_{a,t}$ and $W_{g,t}$ are assigned different values:

- $(W_{a,t} = \bot, W_{g,t} = \bot)$: $\overline{W_{a,t}} \vee W_{g,t}$ is satisfied.

- $(W_{a,t} = \bot, W_{g,t} = \top)$: $\overline{W_{a,t}} \vee W_{g,t}$ is satisfied.

- $(W_{a,t} = \top, W_{g,t} = \top)$: $\overline{W_{a,t}} \vee W_{g,t}$ is satisfied.

- $(W_{a,t} = \top, W_{g,t} = \bot)$: According to Lemma 2, $W_{g,t}$ is a free variable. Therefore, since $\Gamma'_\Psi(W_{g,t}) = \top$, $\overline{W_{a,t}} \vee W_{g,t}$ is satisfied by $\Gamma'_\Psi$, and hence satisfied by $\Gamma_\Pi$.

6. (Transition mutex). Consider any mutex clause between two regular transitions $\delta_1 = \delta_{f \to g}$ and $\delta_2 = \delta_{f' \to g'}$. Let $\delta_{f \to g} \in Trans(a)$ and $\delta_{f' \to g'} \in Trans(b)$, we see that $a$ and $b$ are S-mutex. According to Lemma 3, $a$ and $b$ are also P-mutex in PE. Therefore, we have $\overline{W_{a,t}} \vee \overline{W_{b,t}}$. From our construction, we know $\overline{U_{a,t}} \vee \overline{U_{b,t}}$. Then, since we have the composition of actions, $U_{a,t} \to \bigwedge_{\forall \delta \in Trans(a)} U_{\delta,t}$ and $U_{b,t} \to \bigwedge_{\forall \delta \in Trans(b)} U_{\delta,t}$. A simple resolution of these clauses yields $\overline{U_{\delta_1,t}} \vee \overline{U_{\delta_2,t}}$, which equals to the transition mutex clause $U_{\delta_1,t} \to \overline{U_{\delta_2,t}}$. Therefore, the transition mutex clause is true in $\Gamma_\Pi$. A similar argument applies when the transitions are prevailing and mechanical.

7. (Action existence). The clauses that we want to prove are $U_{\delta,t} \to \bigvee_{\delta \in Trans(a)} U_{a,t}$, for any transitions $\delta$. By our construction, the clauses become

$$\overline{W_{f,t}} \vee \overline{W_{g,t+1}} \vee \bigvee_{\forall a \in A(\delta_{f \to g})} W_{a,t}. \tag{18}$$

Let $\delta = \delta_{f \to g}$. First, we know by definition that $\bigcup_{\forall h} A(\delta_{h \to g}) = \mathrm{ADD}(g)$. Let us denote $\mathrm{ADD}(g)$ as $Z$. According to clauses of class III in PE, there are clauses:

$$W_{g,t} \to \bigvee_{\forall a \in Z} W_{a,t}. \tag{19}$$

We divide $Z$ into multiple action sets according to different fact from $\{f, h_1, \ldots, h_m\}$, denoted as $Z_f, Z_{h_1}, \cdots, Z_{h_m}$. In fact, for each $h \in \{f, h_1, \ldots, h_m\}$, $Z_h$ is equivalent to $A(\delta_{h \to g})$. Consider any $h_i, i = 1, \cdots, m$. According to the clauses of class IV, for every action $a \in \mathrm{PRE}(h)$, there is a clause $W_{a,t} \to W_{h_i,t}$, which is

$$\overline{W_{a,t}} \vee W_{h_i,t}. \tag{20}$$

Next, we perform resolutions by using (19) and all the clauses in (20), for all such $h_i$ and corresponding actions. We consequently have:

$$\overline{W_{g,t}} \vee (W_{h_1,t} \vee W_{h_2,t} \vee \cdots \vee W_{h_m,t}) \vee \bigvee_{\forall a \in Z_f} W_{a,t}. \tag{21}$$

Further, note that all $h_1, h_2, \ldots, f_m$ are mutex to $f$, a resolution using all the mutex clauses in PEresults in:

$$\frac{(21),\ \overline{W_{h_1,t}} \vee \overline{W_{f,t}},\ \overline{W_{h_2,t}} \vee \overline{W_{f,t}},\ \ldots, \overline{W_{h_m,t}} \vee \overline{W_{f,t}}}{\overline{W_{g,t}} \vee (\overline{W_{f,t}} \vee \overline{W_{f,t}} \vee \cdots \vee \overline{W_{f,t}}) \vee \bigvee_{\forall a \in Z_f} W_{a,t}} \tag{22}$$

Since $Z_f = A(\delta_{f \to g})$, the outcome of (22) leads to (18).

8. (Action mutex). Action mutex clauses are satisfied by $\Gamma_\Pi$ according to Lemma 3.

Combining all the cases concludes that the constructed solution $\Gamma_\Pi$ satisfies all the clauses in SASE which means SASE is satisfiable. Since for all action $a$, $\Gamma_\Psi(W_{a,t}) = \Gamma'_\Psi(W_{a,t}) = \Gamma_\Pi(U_{a,t})$, $\Gamma_\Psi$ and $\Gamma_\Pi$ represent the same solution plan. $\qquad\square$

**Theorem 2** *Given a STRIPS task $\Psi$ and a SAS+ task $\Pi$ that are equivalent, for a time step bound $N$, if* SASE$(\Pi, N)$ *is satisfiable,* PE$(\Psi, N)$ *is also satisfiable.*

**Proof** Assuming $\Gamma_\Pi$ is a satisfiable solution to SASE$(\Pi, N)$, we first construct an assignment $\Gamma_\Psi$ from $\Gamma_\Pi$, and show that $\Gamma_\Psi$ satisfies every clause in PE$(\Psi, N)$.

**Construction.** We construct a solution $\Gamma_\Psi$ as follows:

1. For every $a \in \mathcal{A}$ (which is also in $\mathcal{O}$), we let $W_{a,t} = U_{a,t}$;

2. For every dummy action variable $dum_f$, we let $W_{dum_f,t} = U_{\delta_{f\to f},t}$;

3. For every transition $\delta_{f\to g} \in \mathcal{T}$, if $U_{x,f,g,t} = \top$ in $\Gamma_\Pi$, we set $W_{f,t} = W_{g,t+1} = \top$ in $\Gamma_\Pi$;

4. For each fact $f$, if $U_{h,f,t} = \bot$ for every transition $\delta_{h\to f} \in \mathcal{T}$ (which implies that case 3 will not assign a value to $f$), we set $W_{f,t}$ to be $\bot$.

**Satisfiability.** Next, we prove that every clause in PE is satisfied by $\Gamma_\Psi$. The clauses for the initial and goal states are obviously satisfied. Now we consider the add-effect clauses. The clauses that we want to prove are, for every fact $f$:

$$W_{f,t} \to \bigvee_{\forall a \in \text{ADD}(f)} W_{a,t-1} \qquad (23)$$

For a given fact $f$, we consider all the facts $h \neq f$, such that $\delta_{h\to f} \in \mathcal{T}$. For all such $h$, there are two further cases:

- There exists a fact $h$ such that $\delta^x_{h\to f} \in \mathcal{T}$ and $U_{x,h,f,t-1} = \top$ in $\Gamma_\Pi$. In the satisfiable SASE instance, the *action existence* clauses in class F specify that the truth of a non-prevailing transition $\delta$ indicates a disjunction of all actions in $A(\delta)$. Since $U_{x,h,f,t-1} = \top$, it follows that in the SASE instance there is an action $a \in A(\delta^x_{h\to f})$ such that $U_{a,t-1} = \top$. Then, by our construction of $\Gamma_\Psi$, we see that both $W_{h,t-1}$ and $W_{f,t}$ are true. Since $W_{f,t}$ and $W_{a,t-1}$, $a \in \text{ADD}(f)$, are all true, (23) is satisfied by $\Gamma_\Psi$.

- If for every fact $h$ that $\delta^x_{h\to f} \in \mathcal{T}$, $U_{x,h,f,t-1} = \bot$ in $\Gamma_\Pi$, then, according to our construction, $W_{f,t} = \bot$ in $\Gamma_\Psi$. Thus, $\Gamma_\Psi$ satisfies (23).

The two cases above conclude that $\Gamma_\Psi$ satisfies the add effect clauses. Next, we show that $\Gamma_\Psi$ satisfies the precondition clauses, $W_{a,t} \to W_{f,t}$ (i.e. $\overline{W_{a,t}} \vee W_{f,t}$), for all actions $a \in \mathcal{A}$ and facts $f \in pre(a)$. In SASE, we have clauses of class F, which are $U_{a,t} \to U_{\delta,t}$, for all actions $a \in \mathcal{O}$ and $\delta \in Trans(a)$. Let the transition be $\delta^x_{f,g}$, we have $\overline{U_{a,t}} \vee (U_{f,t-1} \wedge U_{g,t-1})$, which implies $\overline{U_{a,t}} \vee U_{f,t-1}$. By our construction, we know $\overline{W_{a,t}} \vee W_{f,t-1}$ is true.

Finally, the mutex clauses are satisfied by $\Gamma_\Psi$ according to Lemma 3. Combining all the cases concludes that the constructed solution $\Gamma_\Psi$ satisfies all the clauses in PE which means PE is satisfiable. $\qquad\square$

## Appendix B. Branching Frequency in All Domains

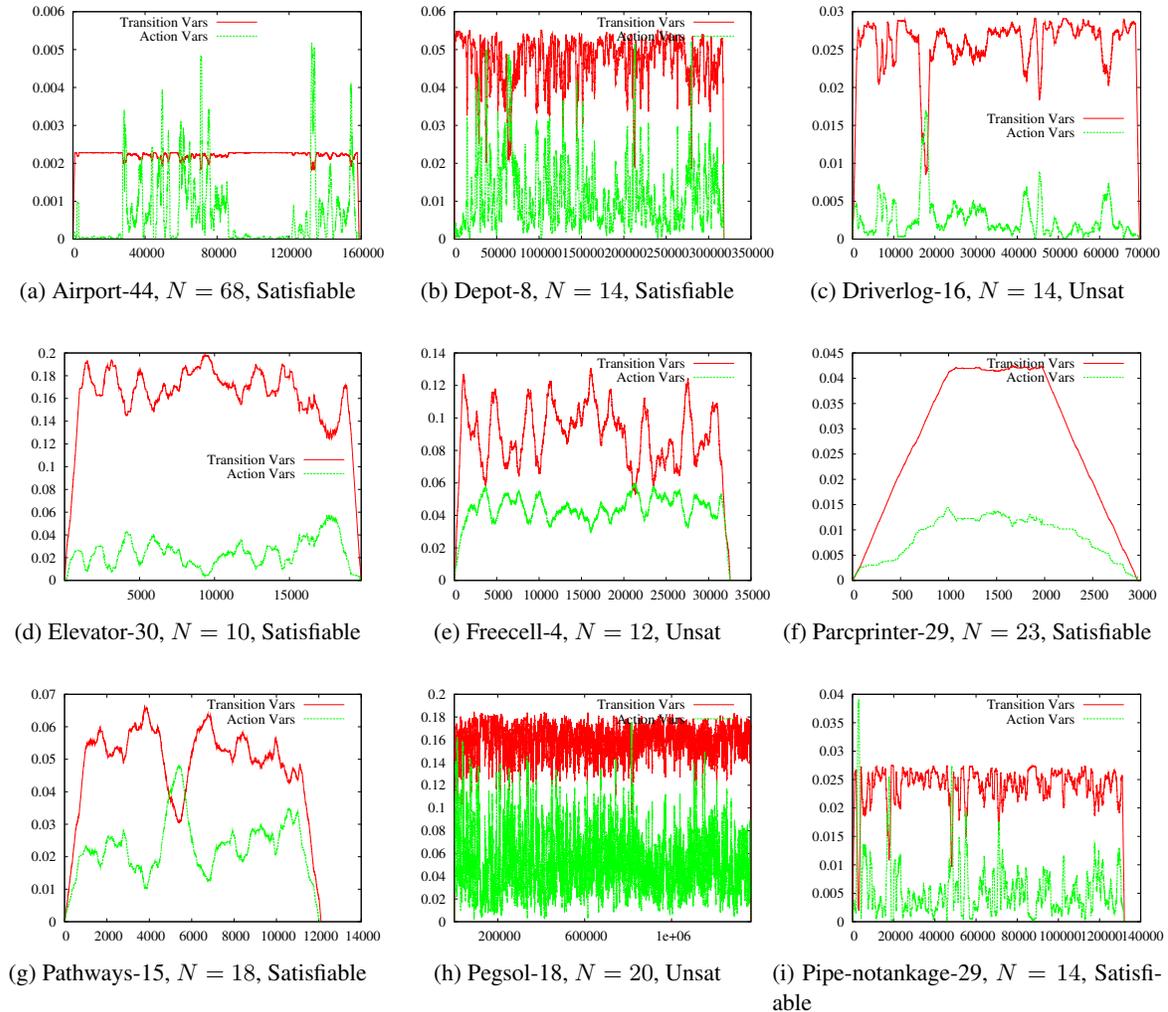

(a) Airport-44, $N = 68$, Satisfiable

(b) Depot-8, $N = 14$, Satisfiable

(c) Driverlog-16, $N = 14$, Unsat

(d) Elevator-30, $N = 10$, Satisfiable

(e) Freecell-4, $N = 12$, Unsat

(f) Parcprinter-29, $N = 23$, Satisfiable

(g) Pathways-15, $N = 18$, Satisfiable

(h) Pegsol-18, $N = 20$, Unsat

(i) Pipe-notankage-29, $N = 14$, Satisfiable

Figure 9: Comparison of variable branching frequency (with $k = 1000$) for transition and action variables in solving certain SAT instances in twelve benchmark domains encoded by SASE. Each figure corresponds to an individual run of MiniSAT. The x axis corresponds to all the decision epochs during SAT solving. The y axis denotes the branching frequency (defined in the text) in an epoch of $k = 1000$.

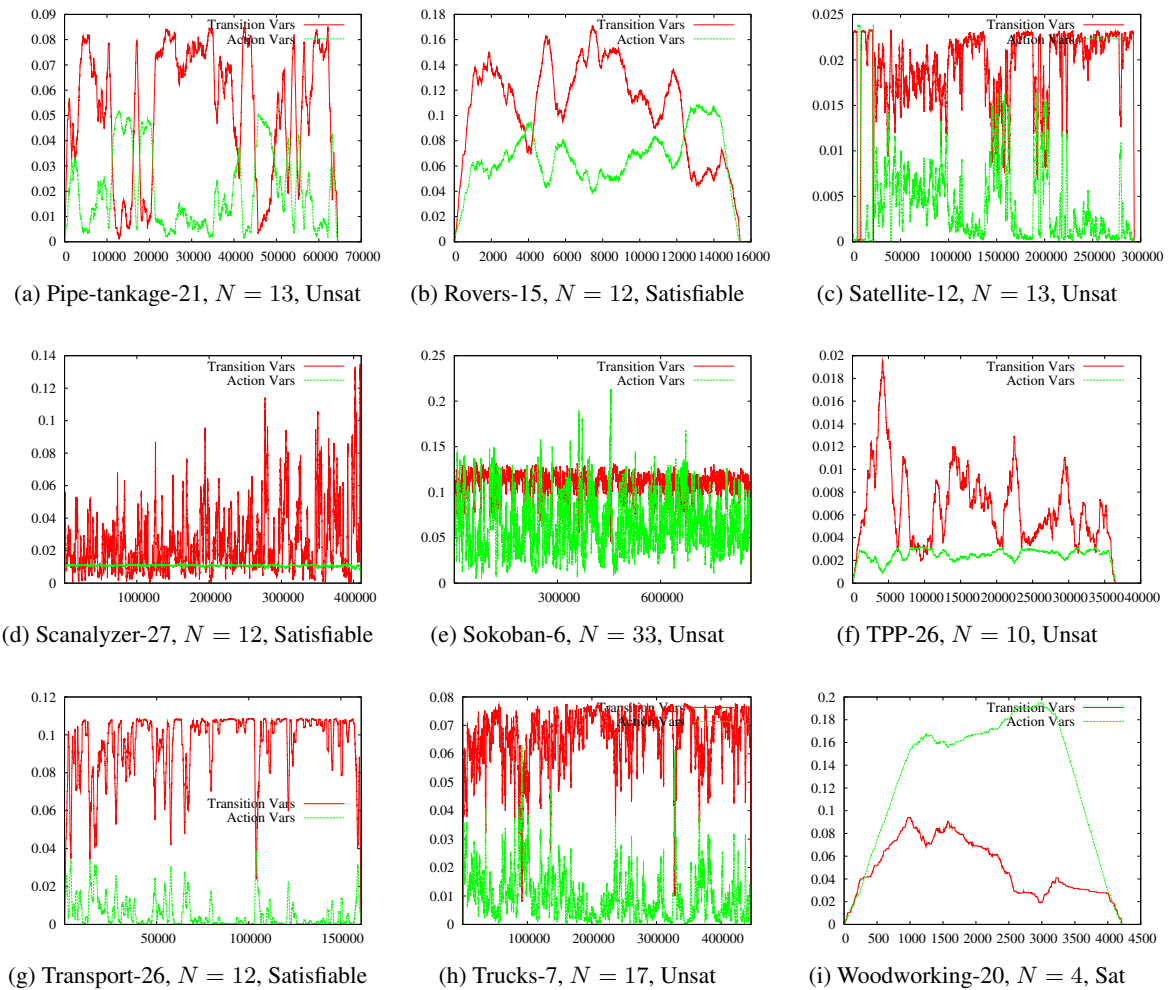

(a) Pipe-tankage-21, $N = 13$, Unsat  (b) Rovers-15, $N = 12$, Satisfiable  (c) Satellite-12, $N = 13$, Unsat

(d) Scanalyzer-27, $N = 12$, Satisfiable  (e) Sokoban-6, $N = 33$, Unsat  (f) TPP-26, $N = 10$, Unsat

(g) Transport-26, $N = 12$, Satisfiable  (h) Trucks-7, $N = 17$, Unsat  (i) Woodworking-20, $N = 4$, Sat

Figure 10: Comparison of variable branching frequency (with $k = 1000$) for transition and action variables in solving certain SAT instances in nine other benchmark domains encoded by SASE.

## References

Bäckström, C., & Nebel, B. (1996). Complexity results for SAS+ planning. *Computational Intelligence*, *11*, 625–655.

Biere, A. (2009). Pr{e,i}coSAT@SC'09. In *SAT'09 Competition*.

Blum, A., & Furst, M. (1997). Fast Planning Through Planning Graph Analysis. *Artificial Intelligence*, *90*, 1636–1642.

Brafman, R. I. (2001). A simplifier for propositional formulas with many binary clauses. In *Proceedings of International Joint Conference on Artificial Intelligence*.

Breiman, L. (1996). Bagging predictors. *Machine Learning*, *24*, 123–140.

Büttner, M., & Rintanen, J. (2005). Satisfiability Planning with Constraints on the Number of Actions. In *Proceedings of International Conference on Automated Planning and Scheduling*.

Castellini, C., Giunchiglia, E., & Tacchella, A. (2003). SAT-based planning in complex domains:Concurrency, constraints and nondeterminism. *Artificial Intelligence*, *147*, 85–117.

Chen, Y., Huang, R., Xing, Z., & Zhang, W. (2009). Long-distance mutual exclusion for planning. *Artificial Intelligence*, *173*, 197–412.

Chen, Y., Huang, R., & Zhang, W. (2008). Fast Planning by Search in Domain Transition Graphs. In *Proceedings of AAAI Conference on Artificial Intelligence*.

Dimopoulos, Y., Nebel, B., & Koehler, J. (1997). Encoding planning problems in nonmonotonic logic programs. In *In Proceeding of the Fourth European Conference on Planning*, pp. 169–181. Springer-Verlag.

Do, B., & Kambhampati, S. (2000). Solving Planning Graph by Compiling it into a CSP. In *Proceedings of International Conference on Automated Planning and Scheduling*.

Ernst, M., Millstein, T., & Weld, D. (1997). Automatic SAT-compilation of planning problems. In *Proceedings of International Joint Conference on Artificial Intelligence*.

Giunchiglia, E., & Maratea, M. (2007). Planning as satisfiability with preferences. In *Proceedings of AAAI Conference on Artificial Intelligence*.

Helmert, M. (2006). The Fast Downward planning system. *Journal of Artificial Intelligence Research*, *26*, 191–246.

Helmert, M. (2008). Concise finite-domain representations for PDDL planning tasks. *Artificial Intelligence*, *173*, 503–535.

Helmert, M., & Domshlak, C. (2009). Landmarks, Critical paths and Abstractions: What's the difference anyway?. In *Proceedings of International Conference on Automated Planning and Scheduling*.

Helmert, M., Haslum, P., & Hoffmann, J. (2008). Explicit-State Abstraction: A New Method for Generating Heuristic Functions. In *Proceedings of AAAI Conference on Artificial Intelligence*.

Hoffmann, J., Gomes, C., & Selman, B. (2006). Structure and Problem Hardness : Goal Asymmetry and DPLL Proofs in SAT-based Planning. In *Proceedings of International Conference on Automated Planning and Scheduling*.

Hoffmann, J., Kautz, H., Gomes, C., & Selman, B. (2007). SAT encodings of state-space reachability problems in numeric domains. In *Proceedings of International Joint Conference on Artificial Intelligence*.

Hoffmann, J., & Nebel, B. (2001). The FF planning system: Fast plan generation through heuristic search. *Journal of Artificial Intelligence Research*, *14*, 253–302.

Huang, R., Chen, Y., & Zhang, W. (2009). An Optimal Temporally Expressive Planner: Initial Results and Application to P2P Network Optimization. In *Proceedings of International Conference on Automated Planning and Scheduling*.

Kautz, H., & Selman, B. (1992). Planning as satisfiability. In *Proceedings of European Conference on Artificial Intelligence*.

Kautz, H., & Selman, B. (1996). Pushing the envelope: Planning, propositional logic, and stochastic search. In *Proceedings of AAAI Conference on Artificial Intelligence*.

Kautz, H., & Selman, B. (1999). Unifying sat-based and graph-based planning. In *Proceedings of International Joint Conference on Artificial Intelligence*.

Kautz, H., Selman, B., & Hoffmann, J. (2006). SatPlan: Planning as Satisfiability. In *5th International Planning Competition, International Conference on Automated Planning and Scheduling*.

Monasson, R., Zecchina, R., Kirkpatrick, S., Selman, B., & Troyansky, L. (1999). Determining computational complexity from characteristic 'phase transitions'. *Nature*, *400(8)*, 133–137.

Moskewicz, M., Madigan, C., Zhao, Y., Zhang, L., & Malik, S. (2001). Chaff: Engineering an Efficient SAT Solver. In *39th Design Automation Conference*.

Myers, J. L., & Well, A. D. (2003). *Research Design and Statistical Analysis* (2nd edition). Routledge.

Nabeshima, H., Soh, T., Inoue, K., & Iwanuma, K. (2006). Lemma reusing for SAT based planning and scheduling. In *Proceedings of International Conference on Automated Planning and Scheduling*.

Ray, K., & Ginsberg, M. L. (2008). The complexity of optimal planning and a more efficient method for finding solutions. In *Proceedings of International Conference on Automated Planning and Scheduling*.

Richter, S., Helmert, M., & Westphal, M. (2008). Landmarks Revisited. In *Proceedings of AAAI Conference on Artificial Intelligence*.

Rintanen, J. (2003). Symmetry Reduction for SAT Representations of Transition System. In *Proceedings of International Conference on Automated Planning and Scheduling*.

Rintanen, J. (2006). Biclique-based representations of binary constraints for making SAT planning applicable to larger problems. In *Proceedings of European Conference on Artificial Intelligence*.

Rintanen, J., Heljanko, K., & Niemelä, I. (2006). Planning as Satisfiability: parallel plans and algorithms for plan search. *Artificial Intelligence*, *12-13*, 1031–1080.

Robinson, N., Gretton, C., Pham, D., & Sattar, A. (2009). SAT-Based Parallel Planning Using a Split Representation of Actions. In *Proceedings of International Conference on Automated Planning and Scheduling*.

Sideris, A., & Dimopoulos, Y. (2010). Constraint propagation in propositional planning. In *Proceedings of International Conference on Automated Planning and Scheduling*.

Soos, M., Nohl, K., & Castelluccia, C. (2009). Extending sat solvers to cryptographic problems. In *International Conference on Theory and Applications of Satisfiability Testing*.

The 6th Int'l Planning Competition (2008). http://ipc.informatik.uni-freiburg.de/homepage/..

The 7th Int'l Planning Competition (2011). http://ipc.icaps-conference.org/..

Williams, R., Gomes, C., & Selman, B. (2003). Backdoors to typical case complexity. In *Proceedings of International Joint Conference on Artificial Intelligence*.

Xing, Z., Chen, Y., & Zhang, W. (2006). MaxPlan: Optimal Planning by Decomposed Satisfiability and Backward Reduction. In *5th International Planning Competition, International Conference on Automated Planning and Scheduling*.

Zhang, W. (2004). Configuration landscape analysis and backbone guided local search: Part I: Satisfiability and maximum satisfiability. *Artificial Intelligence*, *158*, 1–26.

Zhang, W., Rangan, A., & Looks, M. (2003). Backbone Guided Local Search for Maximum Satisfiability. In *Proceedings of International Joint Conference on Artificial Intelligence*.